\pdfoutput=1
\documentclass[numbers]{xoople}

\usepackage{amsfonts}
\usepackage{multirow}
\usepackage{textcomp}
\usepackage{needspace}
\usepackage{placeins}
\usepackage{xspace}

\providecommand{\IEEEPARstart}[2]{#1#2}
\makeatletter
\let\xp@appendix\appendix
\renewcommand{\appendix}[1][]{\xp@appendix
  \section*{Appendix\ifx\relax#1\relax\else: #1\fi}%
  \setcounter{subsection}{0}%
  \renewcommand{\thesubsection}{\Alph{subsection}}%
  \renewcommand{\thesubsubsection}{\thesubsection.\arabic{subsubsection}}}
\makeatother

\makeatletter
\AtBeginEnvironment{table*}{\ifnum\value{table}=0 \fi}
\AtBeginEnvironment{table}{\ifnum\value{table}=5 \fi}
\makeatother

\AtBeginEnvironment{figure*}{\ifnum\value{figure}=6 \setcounter{topnumber}{1}\fi}
\AtBeginEnvironment{table*}{\ifnum\value{table}=8 \setcounter{topnumber}{2}\fi}

\makeatletter
\newif\ifxp@narrow
\AtBeginEnvironment{figure}{\xp@narrowtrue}
\AtBeginEnvironment{table}{\xp@narrowtrue}
\g@addto@macro\@floatboxreset{\ifxp@narrow
  \setlength{\linewidth}{0.6\textwidth}\setlength{\columnwidth}{0.6\textwidth}\centering\fi}
\newsavebox\xp@fitbox
\newif\ifxp@infit
\newif\ifxp@opened
\AtBeginEnvironment{table*}{%
  \let\xp@origtabular\tabular \let\xp@origendtabular\endtabular
  \renewenvironment{tabular}%
    {\ifxp@infit\xp@openedfalse\expandafter\xp@origtabular\else
       \xp@infittrue\xp@openedtrue\begin{lrbox}{\xp@fitbox}\expandafter\xp@origtabular\fi}%
    {\xp@origendtabular
     \ifxp@opened\end{lrbox}%
       \ifdim\wd\xp@fitbox>\linewidth\resizebox{\linewidth}{!}{\usebox\xp@fitbox}%
       \else\usebox\xp@fitbox\fi
     \fi}}
\makeatother

\title{Temporal Sensitivity Analysis of Tessera Embeddings}
\reporttype{Preprint}

\author[*]{Julia Guerrero-Viu}
\author[*]{Alex L\'opez-Cifuentes}
\author{Ignacio P\'erez-Villar}
\author{Fabio Pacifici}
\contribution[*]{Equal contribution}

\date{November 29, 2026}  %
\metadata[document]{Status}{Accepted for publication in the IEEE Journal of Selected Topics in
  Applied Earth Observations and Remote Sensing (JSTARS)}

\footertext{\textcopyright\,2026 The Authors. CC BY 4.0. Accepted for publication in IEEE JSTARS.}

\begin{document}

\begin{abstract}
Many Earth Observation applications need land-use/land-cover maps that are both precise and frequently updated, yet the strongest Earth Observation foundation models build their embeddings from a full year of observations.
We present a controlled study of the temporal sensitivity of Tessera, one of these leading foundation models, for land-use/land-cover mapping. Keeping the encoder frozen, we recompute its embeddings over varying observation windows, from a full year down to a single day. We use them as inputs to a
linear probe and a UNet segmentation head, benchmarking both of them against from-scratch
networks on LUCAS, DynamicEarthNet, and PASTIS-R datasets.
We show that the value of the embeddings
is task-dependent. Where classes are separated by phenology, as for the crop types of
PASTIS-R, they reach a mean Intersection-over-Union of $58.3$, $13\%$ above a temporal network trained from scratch (U-TAE) and $46\%$ above composite-based UNets. Where
classes are temporally stable (e.g., forests in DynamicEarthNet and LUCAS),
embedding-based and from-scratch models match only under full supervision. On both
datasets, Tessera embeddings remain markedly more label-efficient.
Degradation under shorter temporal windows is gradual and class-dependent. Contracting the window
from one year to one month costs $39\%$ of the segmentation accuracy on PASTIS-R but only
$5\%$ on DynamicEarthNet. Single-day embeddings still classify land cover in LUCAS with a balanced accuracy $2.9$ times the chance level.
Our study shows that temporal coverage is therefore a tunable cost
rather than a fixed prerequisite, opening regimes such as low-latency mapping and faster land-use/land-cover refresh cycles.
\end{abstract}
\maketitle

\section{Introduction}
\label{sec:intro}
\IEEEPARstart{A}{ccurate} and up-to-date land-use/land-cover (LULC) maps are a core product of Earth Observation (EO), essential for agricultural monitoring, biodiversity assessment, semantic change detection, land-use policy and disaster response. How frequently such maps must be refreshed, however, depends strongly on the application: urban expansion or forest extent are adequately tracked with an annual update, whereas monitoring crops within a growing season, or assessing damage after a flood or a fire, requires far shorter latency. Today's operational products make this gap concrete. The most widely used global land-cover maps, such as ESA WorldCover~\cite{zanaga2022worldcover} and NASA's MODIS land-cover product~\cite{friedl2022mcd12q1}, are refreshed at most once a year and released months after the mapped period, so their effective update latency is on the order of a year. Faster, near-real-time products such as Dynamic World~\cite{brown2022dynamicworld} do exist, but shorter latency generally comes at the cost of map accuracy or spatial resolution.
Producing maps at high spatial resolution \emph{and} short latency remains challenging, as it must cope with frequent cloud contamination of the optical acquisitions, sparse in-situ annotations, and the large volume of multi-temporal satellite data.

Two of these challenges, scarce annotations and the sheer volume of unlabeled imagery, are exactly what the foundation model (FM) paradigm turns into an advantage, pretraining once on vast unlabeled data and reusing the representation across tasks with little supervision~\cite{bommasani2021foundation}. Having reshaped language and vision~\cite{devlin2019bert,he2022mae,radford2021clip,kirillov2023sam}, the recipe has reached EO, whose archives are vast and almost entirely unlabeled while in-situ annotations are costly and scarce. EO FMs such as Prithvi-EO-2.0~\cite{szwarcman2025prithvi}, AlphaEarth~\cite{brown2025alphaearth} and Tessera~\cite{feng2025tessera} now produce general-purpose embeddings that transfer across many downstream tasks, and those that exploit long temporal sequences outperform monotemporal ones.

In particular, both AlphaEarth~\cite{brown2025alphaearth} and Tessera~\cite{feng2025tessera} follow the same recipe: encoding a full year of multi-modal satellite time series into a dense per-pixel embedding at $10$-meter resolution. We focus on Tessera, whose strictly per-pixel design applies temporal attention directly to each pixel's own Sentinel-1 Radiometric Terrain Corrected (RTC) and Sentinel-2 spectral series. This granularity matches the per-pixel nature of land-cover mapping, and its attention over each pixel's full spectral-temporal profile is well suited to the phenological differences that separate many land-cover classes.
Therefore, its pixel-wise representation yields higher-quality embeddings than ViT-based monotemporal EO FMs (e.g., DINOv3-SAT~\cite{simeoni2025dinov3}), which operate on coarser, patch-level features. That gain, however, comes at a price: a full year of observations, precisely the coverage that the low-latency, high-resolution mapping demanded by crop monitoring or disaster response cannot wait for. While Tessera reports robustness to which observations are sampled within the year and to removing one season from it~\cite{feng2025tessera}, and AlphaEarth supports partial-year inference by design, neither provides a systematic, controlled evaluation of how downstream performance degrades as the observation window from which the frozen encoder computes its embeddings is shortened. We therefore ask:
\begin{quote}
\centering
\emph{\textbf{How robust are frozen Tessera embeddings to reductions in the temporal observation window over which they are computed?}}
\end{quote}

To address it, we present a systematic study of the temporal sensitivity of Tessera embeddings for land-cover classification and segmentation with the following contributions:
\begin{itemize}
\interlinepenalty=10000
    \item We propose a controlled evaluation framework for the temporal sensitivity of a pixel-wise EO foundation model. Keeping the encoder frozen, we recompute its embeddings over observation windows ranging from a full year down to a single day, while the downstream architecture, training recipe and data splits stay fixed. Shortening the window changes the temporal span, the seasonal coverage and the number of observations together, and we report their combined effect, which we then separate on PASTIS-R. We apply the framework to three LULC benchmarks: LUCAS, PASTIS-R, and DynamicEarthNet.
    \item We show that the value of the embeddings is \emph{task-dependent}: at full supervision they are decisive when classes are separated by phenology and only marginal when they are temporally stable. The same dichotomy also reappears within each benchmark at the class level.
    \item We show that the embeddings are markedly more \emph{label-efficient} than training from scratch, retaining a clear advantage in the low-label regime. This matters precisely where annotation cannot be produced in time, as in disaster response and other low-latency settings.
    \item We show that meaningful semantic information is preserved under extreme temporal sparsity, down to single-day windows and well above chance level, enabling operating regimes with drastically reduced temporal coverage such as low-latency mapping and faster LULC refresh cycles.
\end{itemize}

\section{Related Work}
\label{sec:rw}
\subsection{Foundation Models for Earth Observation}
Self-supervised pretraining on large, unlabeled satellite archives has produced
general-purpose encoders that transfer across many downstream tasks. A first family
operates \emph{image-wise}: SatMAE~\cite{cong2022satmae} and
Prithvi-EO-2.0~\cite{szwarcman2025prithvi} pretrain ViT backbones on multispectral
scenes, DOFA~\cite{xiong2024DOFA} conditions a single backbone on the wavelength of
each input channel so that one encoder serves sensors with different band
configurations, and general-purpose vision models such as
DINOv3~\cite{simeoni2025dinov3} have been transferred to EO by re-running generic
self-supervised recipes on Sentinel archives~\cite{wang2023ssl4eo} or by adapting
vision-language and promptable backbones~\cite{remoteclip,wang2023samrs}. Several of
them ingest more than one date, but time enters as an extra token axis of a
\emph{patch} representation, so the output remains a patch-level feature rather than
a per-pixel temporal descriptor. A second family represents each location by its own
\emph{time series}: Presto~\cite{tseng2023presto} learns from multi-sensor pixel
sequences and Galileo~\cite{tseng2025galileolearningglobal} from multi-scale
space--time blocks, while AlphaEarth~\cite{brown2025alphaearth} and
Tessera~\cite{feng2025tessera} produce dense annual embeddings at native
($10\,\mathrm{m}$) resolution; AlphaEarth fuses ten gridded sources plus text into a
$64$-dimensional embedding field with spatial self-attention, whereas Tessera is
strictly per-pixel and uses only Sentinel-1 and Sentinel-2. Our study concerns this
temporal family: rather than proposing a new encoder, we characterize how the
embeddings of a temporally-aggregated model behave when the temporal evidence
available at inference is restricted.

\emph{Tessera}~\cite{feng2025tessera}, the foundation model our analysis builds on,
encodes each $10\,\mathrm{m}$ pixel's full annual Sentinel-1/2 time series into a single
$128$-dimensional embedding: two four-block transformer encoders, one per modality
(ten Sentinel-2 bands and Sentinel-1 VV/VH), consume the sequences with day-of-year
positional encodings and attention pooling, and are fused by an MLP. Pretraining
follows a Barlow-Twins objective over two views of the same pixel-year, each a sparse
random sample of $40$ valid acquisitions, on $\sim$$0.8$ billion pixels drawn from
$3{,}012$ globally distributed tiles ($2017$--$2024$), and the resulting frozen
embeddings prove strongly label-efficient across classification, segmentation and
regression tasks when read out by lightweight heads. Its self-supervision, however,
enforces invariance only to which observations within the year are sampled, and both pretraining and inference assume a full-year temporal support. The authors note that shorter seasonal windows at inference, such as spring to autumn, still yield useful embeddings, and in an ablation on a single crop-type dataset (Austrian Crop) removing one season from the annual series changes F1 by less than one point~\cite{feng2025tessera}. Shorter contiguous windows, other mapping tasks and the effect on individual classes are not analyzed.

\subsection{Temporal Modeling of Satellite Image Time Series}
The temporal dimension is decisive for many EO tasks, most notably crop-type mapping,
where classes are separated by \emph{phenology} rather than by any single-date
spectral signature~\cite{pelletier2019temporal}. This has motivated a long line of
task-specific architectures for satellite image time series, from recurrent
encoders~\cite{russwurm2018multi} and temporal convolutions~\cite{pelletier2019temporal}
to temporal-attention models such as U-TAE~\cite{garnot2021panoptic} and the
fully-attentional TSViT~\cite{tarasiou2023tsvit}, with multimodal optical–radar fusion
further improving accuracy and cloud resilience~\cite{garnot2022pastisR}. These models
are trained end-to-end for each task and region, whereas foundation models learn the
temporal representation once and reuse it across tasks. Whether such a generic,
temporally-aggregated embedding retains the phenological information that
task-specific models exploit is therefore task-dependent, and is a central question of
our analysis on phenology-driven versus spectrally-stable
benchmarks~\cite{garnot2022pastisR,toker2022dynamicearthnet}.

\subsection{Land Cover Classification Benchmarks}
Land-cover evaluation resources are abundant, ranging from patch-level scene
classification~\cite{helber2019eurosat,sumbul2019bigearthnet} and single-date dense
segmentation~\cite{wang2021loveda} to global map products~\cite{brown2022dynamicworld},
crop-oriented time-series collections~\cite{russwurm2020breizhcrops,schneider2023eurocrops}
and the benchmarks created to standardize foundation-model evaluation,
GEO-Bench~\cite{lacoste2023geobench} and PANGAEA~\cite{marsocci2024pangaea}. For most
of the datasets in these benchmarks, however, neither the temporal axis nor the full multispectral input is
included: the majority are mono-temporal, and several provide only RGB or a reduced band
set. Only a few, such as PASTIS-R, expose a genuine Sentinel-1/2 time series, and
DEN itself enters PANGAEA as mono-temporal RGB Planet imagery rather than as
a multispectral series, so in these configurations the observation window cannot be
varied at all. Our temporal sensitivity analysis admits only benchmarks whose labels
carry a reference date to anchor the window, are
pixel-wise on the $10\,\mathrm{m}$ grid and come from human reference data rather than
from another model's predictions, and for which the raw Sentinel-1/2 acquisitions can
be re-encoded at every window length. Reliable ground truth is also crucial because we
measure the information content of a representation rather than build a mapping
product, so inherited labels would confound the degradation we attribute to a shortened
window with the labelling model's own errors. Among the few resources that qualify, we
retain the three detailed in Section~\ref{sec:datasets} (LUCAS, DEN and PASTIS-R),
which jointly span annotation
cadence, label density and phenological dependence.
To the best of our knowledge, existing LULC benchmarks in the literature
use a fixed temporal extent at evaluation time: every model is scored on the complete record,
be it a full annual composite or an entire acquisition series. The extent of the
observation window is thus a constant of the evaluation protocol rather than one of its
variables, so previous results report a single operating point and do not analyze how
performance evolves as the available coverage shrinks. Turning that constant into a
controlled variable is precisely what our novel evaluation framework allows.

\section{Methodology}
\label{sec:method}

We investigate how the \emph{temporal observation window} shapes the embeddings of Tessera model for land-cover mapping.
In this section, we first describe the three datasets used and their label pre-processing (Sec.~\ref{sec:datasets}); then, we formalize the proposed temporal sensitivity evaluation framework (Sec.~\ref{sec:framework}); finally, we explain the classification and segmentation heads (Sec.~\ref{sec:probe}).

\subsection{Datasets}
\label{sec:datasets}

Our analysis includes three public benchmarks chosen to span complementary evaluation regimes and taxonomies: sparse in-situ polygons (LUCAS), and dense multi-temporal rasters for two different tasks, general land-cover segmentation (DEN) and phenologically-driven crop-type mapping (PASTIS-R).
In all cases, labels are preprocessed onto the native $10\,\mathrm{m}$ Tessera embedding grid so that a probe can be trained and scored per pixel (see the Appendix for details on data acquisition and preprocessing).

\subsubsection{LUCAS Dataset}
LUCAS 2022~\cite{dandrimont2022LUCASdata} is a pan-European in-situ land-cover survey, from which we draw a stratified subset of $15{,}000$ georeferenced polygons and pair them with Copernicus Sentinel-1/2 imagery encoded by Tessera~\cite{feng2025tessera}. We adopt the \emph{level-1} LC1 taxonomy (the leading character of the land-cover code, e.g.\ \texttt{C30}$\rightarrow$\texttt{C}), giving eight top-level classes: artificial land, cropland, woodland, shrubland, grassland, bare land, water, and wetland. Each polygon carries a precise survey date, making LUCAS well suited to our controlled temporal analysis, and we rasterize them onto the $10\,\mathrm{m}$ grid to match the pixel-wise embeddings. The survey is spatially precise (in-situ validated by experts) but highly sparse (isolated polygons rather than dense coverage) and strongly class-imbalanced, being dominated by cropland, grassland and woodland; moreover, many polygons are smaller than the $10\,\mathrm{m}$ Tessera grid, so individual samples are prone to mixed-pixel effects. As there is no standard train/test split available for this benchmark, we follow previous work~\cite{brown2025alphaearth} and create a balanced train split to mitigate class imbalance. Our training set contains 100 polygons per class, reserving the remaining samples for evaluation. The training polygons are drawn at random within each class, without spatial constraints. Because LUCAS points lie on a regular grid, no test polygon is closer than $1.7$\,km to a training polygon (median distance $32.8$\,km), but $95\%$ of the test polygons lie in a NUTS-2 region that also contains training polygons, so the evaluation is not spatially independent at the regional scale.

\subsubsection{DynamicEarthNet Dataset (DEN)}
DEN~\cite{toker2022dynamicearthnet} targets dense land-cover mapping over $75$ areas of interest distributed worldwide (of which we use $50$ for training and $20$ for testing, see Section~\ref{subsubsec:training_setup}), each a $1024{\times}1024$ tile imaged daily by $3\,\mathrm{m}$ Planet mosaics across two years ($2018$--$2019$) and annotated per pixel at \emph{monthly} cadence with seven classes: impervious surface, agriculture, forest and other vegetation, wetlands, soil, water, and snow/ice. We encode the corresponding Sentinel-1/2 time series with Tessera on the $10\,\mathrm{m}$ grid and therefore aggregate the $3\,\mathrm{m}$ labels to $10\,\mathrm{m}$,
yielding monthly per-pixel labels $(\mathrm{scene},\mathrm{row},\mathrm{col},y)$ aligned with the embeddings under the standard DEN split. Like LUCAS, DEN is strongly imbalanced: the forest/vegetation and soil classes dominate, whereas wetlands are scarce and snow/ice is absent in the official test split.

\subsubsection{PASTIS-R Dataset}
PASTIS-R~\cite{garnot2022pastisR} addresses crop-type mapping over $2{,}433$ patches of $128{\times}128$ pixels at $10\,\mathrm{m}$, sampled across four Sentinel-2 tiles in metropolitan France and observed over a single agricultural year (Sept.\ $2018$--Nov.\ $2019$). Each patch provides co-registered Sentinel-1/2 time series and a single time-invariant annotation of $18$ crop types plus a background class ($19$ scored classes; the \emph{void}/no-data label is ignored). Its labels are natively at $10\,\mathrm{m}$, so we align each patch's target raster to the embedding grid, and we use the official split (i.e., folds 1--3 train, fold 4 validation, fold 5 test). PASTIS-R is likewise highly imbalanced, dominated by the background class and meadow, with several crop types (e.g.\ sorghum, mixed cereal) only sparsely represented.

\subsection{Temporal Sensitivity Evaluation Framework}
\label{sec:framework}

\begin{figure*}[!t]
\centering
\includegraphics[width=\textwidth]{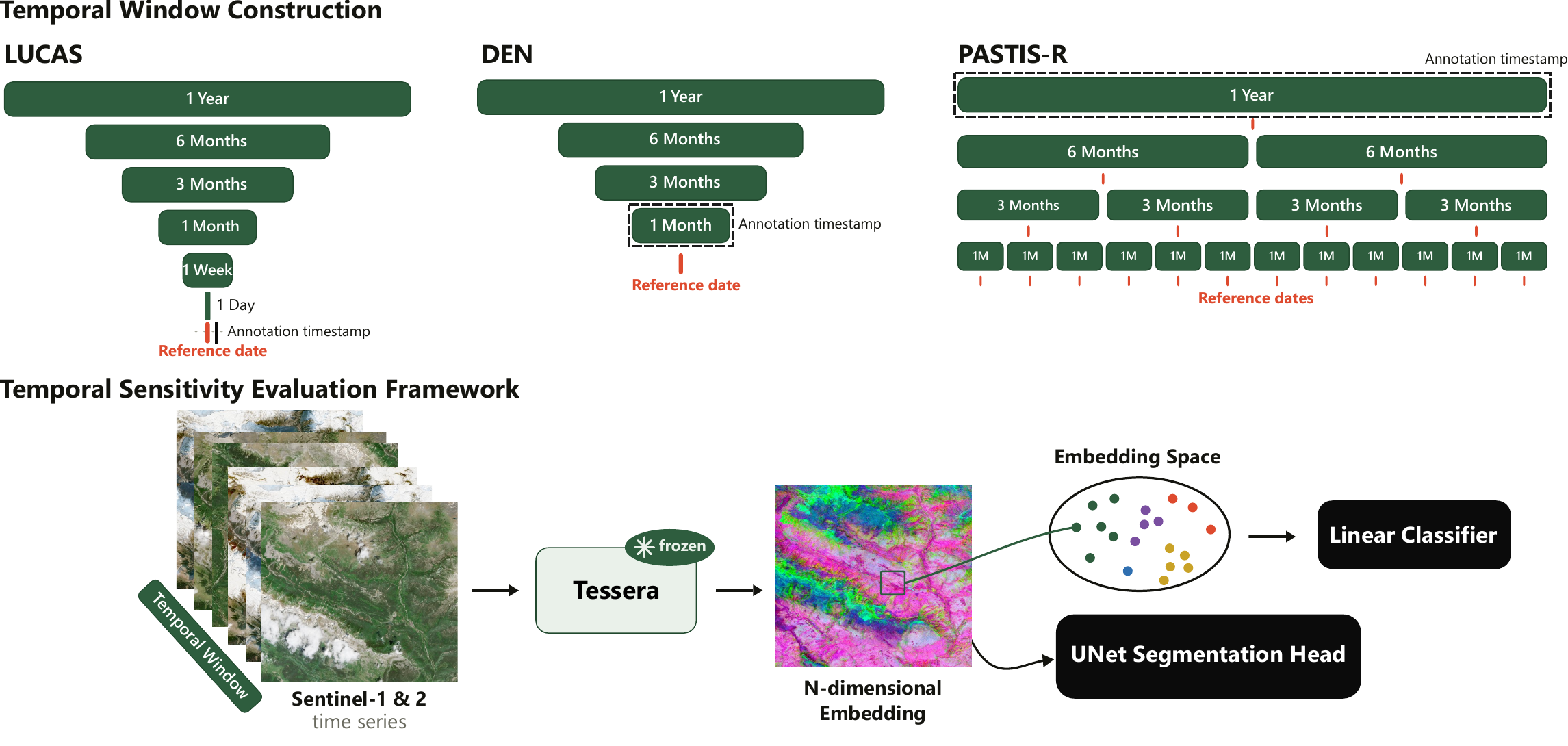}
\caption{Temporal Sensitivity Evaluation Framework: (Top) For each dataset, multi-temporal Sentinel-1/-2 observations are extracted over observation windows of decreasing length defined relative to a dataset-specific reference date (red), which differs from the annotation timestamp (black) as described in Section~\ref{sec:framework}. (Bottom) Given all observations in a temporal window, we encode them with the frozen Tessera model into a dense per-pixel embedding map. For each dataset and temporal window, we train a linear classification probe and a UNet segmentation head, with the same architecture and training settings at every window, to evaluate the quality of the Tessera embeddings at that temporal resolution.}
\label{fig:windows_construction}
\end{figure*}

As already introduced, original Tessera encodes each pixel's full \emph{one-year} Sentinel-1/2 time series into a single embedding: it exposes no notion of a variable observation window. We propose to generalize this to a controllable \emph{temporal receptive field}, so that the influence of temporal support on representation quality can be measured directly, and compared with the one-year setting.
Figure~\ref{fig:windows_construction} illustrates the framework, which keeps the frozen encoder, the architecture and training settings of the downstream heads, and the data splits fixed, and varies the extent of the temporal observation window over which per-pixel embeddings are computed and evaluated. The parameters of the encoder never change, whereas a new head is fitted for each window (Section~\ref{sec:probe}).

\subsubsection{Temporal Window Construction}
Let $\mathcal{T}=\mathbb{R}$ denote calendar time measured in days. For a pixel $x$ and sensor $m\in\{\mathrm{S1},\mathrm{S2}\}$, let its Sentinel time series be the set of \emph{valid} acquisitions
\begin{equation}
  \mathcal{S}^m(x)\subset \mathbb{R}^{C_m}\times\mathcal{T},
\end{equation}
whose elements are acquisition pairs $(\mathbf{b},\tau)$, where $\mathbf{b}\in\mathbb{R}^{C_m}$ is the multi-channel observation vector of sensor $m$, with $C_{\mathrm{S1}}=2$ radar backscatter channels from Sentinel-1 RTC and $C_{\mathrm{S2}}=10$ optical reflectance bands from Sentinel-2 (cloud-screened), and $\tau \in\mathcal{T}$ is its calendar timestamp.
Let $f_\theta$ denote the frozen Tessera encoder, which maps such a time series to a $d$-dimensional embedding ($d{=}128$). Importantly, the Tessera encoder $f_\theta$ operates on a \emph{fixed-length} input so this length is a property of the model and is different from the observation window that we define next.

We parameterize the observation window by its temporal extent $W$ in days (e.g., 30 for one month, or 365 for one year), together with a reference date $t$ around which the window is centered:
\begin{equation}
  \Delta_W(t)=[t-W/2, t+W/2] \subset \mathcal{T}, \quad W > 0.
\end{equation}
Since the window is centered, the embedding for a reference date $t$ uses acquisitions up to $W/2$ days after $t$. In an operational setting, a map for date $t$ would therefore be available with a delay of $W/2$ (about $15$ days for 1M), which remains far shorter than the delay of annual products but is not real-time; trailing windows $[t-W,t]$ would remove this delay and are not evaluated here.

The composite embedding is then obtained by restricting the encoder's input to the acquisitions whose timestamps fall inside this window. Let
\begin{equation}
  \mathcal{A}^m_W(x,t)=\{(\mathbf{b},\tau)\in \mathcal{S}^m(x):\tau \in\Delta_W(t)\}
\end{equation}
be the set of in-window acquisitions of sensor $m$. Since $f_\theta$ always ingests exactly a fixed length of $n$ observations, its input is formed by drawing $n$ samples from $\mathcal{A}^m_W(x,t)$ (sub-sampling when more than $n$ are available and \emph{repeating} acquisitions when fewer, matching the original setup in Tessera; the $n$ indices are equally spaced over the in-window acquisitions, so the sampling is deterministic). The encoder has no attention mask and always receives $n$ steps per sensor, so a sensor without any acquisition in the window cannot simply be omitted. On LUCAS, a missing Sentinel-1 sensor is filled with zeros, which the encoder sees as a constant input; Sentinel-2 is never missing, because $t$ is a Sentinel-2 acquisition date. Sentinel-1 is missing in $75\%$ of the LUCAS test samples at 1D and $3\%$ at 1W, and never at longer windows, so the 1D embeddings rest essentially on a single Sentinel-2 date. On DEN and PASTIS-R, a sensor without acquisitions in the window is replaced by its closest acquisition outside the window, repeated $n$ times. This affects $18.8\%$ (Sentinel-2) and $3.8\%$ (Sentinel-1) of the DEN test samples at 1M, $6.7\%$ and $2.3\%$ at 3M, and $1.5\%$ and $0.8\%$ at 6M, and $5.9\%$ of the PASTIS-R test windows at 1M (Sentinel-2 only), with the replacement acquisition at most $143$ days from $t$ on DEN and $42$ on PASTIS-R. Pixels without any non-zero Sentinel-2 value receive a zero embedding. Appendix~\ref{app:observations} reports the number of observations per window. Writing $\Pi_{n}$ for this fixed-size sampling, applied independently to each sensor, the composite embedding is
\begin{equation}
  \mathbf{e}_W(x,t)=f_\theta\big(\Pi_{n}\!\left(\mathcal{A}^{S1}_W(x,t)\right),\Pi_{n}\!\left(\mathcal{A}^{S2}_W(x,t)\right)\big)\in\mathbb{R}^{d}.
\end{equation}
Varying $W$ therefore changes the number of distinct acquisitions $|\mathcal{A}^m_W(x,t)|$, the time span over which they are spread and the part of the year that is observed, and setting $W{=}\mathrm{1Y}$ recovers the original model. These three factors vary together, so the effect we report for a given window length is their combined effect under the observation conditions of each benchmark.
In our study, due to the Tessera model design, the length of observations ingested by the model is fixed at $n=40$ for all window lengths, and the embedding space is $d=128$-dimensional.

\subsubsection{Reference Dates}
In all benchmarks, for an annotation with timestamp $t_{ann}$ (the date to which the label refers), we take the reference date $t$ as the Sentinel-2 acquisition date of $x$ closest to $t_{ann}$ on LUCAS (median offset $3$ days, maximum $101$), and as the 15th day of the labeled month on DEN, whose labels are monthly rasters dated on the first day and center the observation window on it. All windows of a sample are centered on the same reference date, which therefore differs from $t_{ann}$ (Figure~\ref{fig:windows_construction}, annotation timestamp in black and reference date in red); on LUCAS, the 1Y window is the calendar year 2022.
Then, for each benchmark, we define the set of reference dates based on the availability and resolution of their annotations, as follows:
\textbf{LUCAS} and \textbf{DEN} provide one label per sample, with annotation timestamps $t_{ann}$ at two different resolutions, daily per LUCAS and monthly for DEN. \textbf{PASTIS-R}, in contrast, provides a single annual label per patch; we therefore tile the year from 1~October~2018 to 30~September~2019 into non-overlapping windows of length $W$ ($12$, $4$, $2$ and $1$ per patch for 1M, 3M, 6M and 1Y) with the centered reference dates, so that the year-long label supervises each composite spanning that year, as illustrated in Figure~\ref{fig:windows_construction}. On PASTIS-R, short windows therefore measure how well part of the season identifies the crop of the whole season, rather than the state of the parcel within that window. We study $W\in\{\mathrm{1M},\mathrm{3M},\mathrm{6M},\mathrm{1Y}\}$ on the dense benchmarks (DEN and PASTIS-R), and additionally include one-week and one-day windows for LUCAS for completeness $W\in\{\mathrm{1D},\mathrm{1W},\mathrm{1M},\mathrm{3M},\mathrm{6M},\mathrm{1Y}\}$, as this benchmark includes precise daily timestamps per label.

\subsection{Classification and Segmentation Heads}
\label{sec:probe}

Both heads are trained anew for each pair of benchmark and window $W$, with identical architecture, hyperparameters and data splits, and evaluated on embeddings computed over the same window. This is the standard protocol for evaluating frozen representations: the head that can be fitted on the embeddings of a window measures how much task-relevant information they contain. It does not measure the robustness of a fixed downstream model applied to windows other than the one it was trained on, which we discuss in Section~\ref{sec:discussion}.

\subsubsection{Linear Probe Head}
To measure the \emph{intrinsic} quality of the embeddings, and to attribute the observed differences to the observation window $W$ rather than to the decoder, we first attach the simplest possible decoder, a linear classifier, and deliberately employ no temporal or spatial post-processing.
For each (benchmark, $W$) pair we $\ell_2$-normalize the embeddings and fit a logistic-regression probe with inverse-regularization strength $\lambda{=}10$ and class weights $w_c\propto 1/\sqrt{\mathrm{freq}(c)}$, which mitigate the class imbalance of the dense benchmarks; on LUCAS, whose training split is balanced, they are uniform. Both settings are fixed a priori and not tuned, and are the same for every benchmark and window.
Because the dense benchmarks (DEN and PASTIS-R) contain millions of labeled pixels, for faster convergence the probe is trained on a $10\%$ subsample of the training set that preserves the class distribution, and the fit is repeated on $5$ independently drawn subsamples (Section~\ref{subsubsec:training_setup}). In the test set, we apply the probe to every pixel and produce dense prediction maps, so that evaluation metrics remain comparable to dense-prediction baselines.

\subsubsection{UNet Head}
For the task of dense segmentation (DEN and PASTIS datasets), however, a linear probe has limited capacity and assumes pixel independency. Therefore, and following Tessera~\cite{feng2025tessera}, we extend the analysis by decoding the frozen, precomputed Tessera embeddings with a UNet-based segmentation head.
The head is a faithful reimplementation of the compact DoubleConv UNet used in the Tessera downstream-task code\footnote{\url{https://github.com/ucam-eo/tessera}}. We reproduce Tessera's training recipe as faithfully as the details reported in the original paper allow, and train it on PASTIS-R and DEN. Since some training details are not fully specified, and to guard against undertraining, we validate our design choices with the ablations reported in Section~\ref{subsubsec:ablation_studies} (Tables~\ref{tab:ablation_DEN} and~\ref{tab:ablation_PASTIS}).

\definecolor{firstbest}{RGB}{0,140,60}
\definecolor{secondbest}{RGB}{0,90,200}
\definecolor{gainpos}{RGB}{0,140,60}
\definecolor{lossneg}{RGB}{200,30,30}

\newcommand{\fst}[1]{\textcolor{firstbest}{\textbf{#1}}}
\newcommand{\snd}[1]{\textcolor{secondbest}{\textbf{#1}}}
\newcommand{\pms}[1]{\,\raisebox{0.35ex}{\tiny$\pm#1$}}
\newcommand{\nps}{\hphantom{\,\raisebox{0.35ex}{\tiny$\pm0.0$}}}
\newcommand{\dpos}[1]{\textcolor{gainpos}{+#1}}
\newcommand{\dneg}[1]{\textcolor{red}{$-#1$}}

\section{Experimental Results}
\label{sec:results}
This section evaluates the proposed methodology on the datasets described in Section~\ref{sec:method}. Section~\ref{subsec:implementation_details} covers the implementation details common to all evaluations, Section~\ref{subsec:quantitative_results} presents the main quantitative results and ablation studies, Section~\ref{subsubsec:temporal_factors} separates the factors behind the temporal sensitivity, and Section~\ref{subsec:qualitative_results} presents qualitative results.

\subsection{Implementation Details}
\label{subsec:implementation_details}
The construction of the temporal windows, the train/validation/test splits, and the linear-probing configuration are part of the core evaluation framework and are described in Section~\ref{sec:method}. This section details the remaining implementation aspects: the training setup shared across experiments (Section~\ref{subsubsec:training_setup}), the baseline models used for comparison (Section~\ref{subsubsec:compared_models}), and the evaluation metrics (Section~\ref{subsubsec:evaluation_metrics}).

\subsubsection{Training Setup}
\label{subsubsec:training_setup}
All trained segmentation models share an identical training recipe, so that any difference between them stems solely from the input rather than from the training procedure. We use the Adam optimizer (learning rate $10^{-4}$, weight decay $10^{-4}$) with a cosine-annealing schedule over $80$ epochs, and a batch size of $64$ for PASTIS-R ($128{\times}128$ crops) and $96$ for DEN ($256{\times}256$ crops).

The objective is a class-weighted pixel-wise cross-entropy that ignores void/unlabeled pixels,
\begin{equation}
  \mathcal{L} = -\frac{\displaystyle\sum_{i \in \Omega} w_{y_i} \, \log \frac{\exp(z_{i,y_i})}{\sum_{c\in\mathcal{C}} \exp(z_{i,c})}}{\displaystyle\sum_{i \in \Omega} w_{y_i}} ,
\end{equation}
where $\Omega$ is the set of labeled pixels, $\mathcal{C}$ the set of classes, $y_i$ the ground-truth class of pixel $i$, $z_{i,c}$ the predicted logit for class $c$, and the normalization by $\sum_{i} w_{y_i}$ follows the weighted-mean reduction of the loss. To mitigate class imbalance, the per-class weight is the inverse square root of the class frequency,
\begin{equation}
  \tilde{w}_c =
  \begin{cases}
    \dfrac{1}{\sqrt{p_c}}, & N_c > 0 \\[4pt]
    0,          & N_c = 0
  \end{cases} ,
  \qquad
  w_c = \frac{|\mathcal{C}^{+}|}{\sum_{c'} \tilde{w}_{c'}} \, \tilde{w}_c ,
\end{equation}
with $p_c = N_c / \sum_{c'} N_{c'}$ the frequency of class $c$, $N_c$ the number of training pixels of class $c$, and $\mathcal{C}^{+}$ the set of classes present in the training split ($N_c>0$). The weights are normalized so that those of the present classes have mean one (equivalently, they sum to $|\mathcal{C}^{+}|$); classes absent from the split receive zero weight.

\textbf{Preprocessing and initialization.} Input embeddings are per-channel standardized using training-set statistics. All trainable weights are randomly initialized with Kaiming (He) uniform initialization for the convolutional and linear layers. Training uses mixed precision, and the same autocast context wraps the forward pass at evaluation time.

\textbf{Model selection.} For every trained model we report the checkpoint with the highest validation mIoU over the $80$ epochs. On PASTIS-R the validation split is the official fold~4. DEN has no validation split, so we divide its $50$ training areas into five folds of $10$ areas each: the run with seed $s$ is validated on fold $s$ and trained on the other four, so that every run is trained on $80\%$ of the training areas. The test split ($20$ areas, $480$ monthly samples) is used neither for training nor for model selection. The same selection is used for the configuration ablation of Section~\ref{subsubsec:ablation_studies}.

\textbf{Seeds and reporting.} Every configuration is repeated $5$ times and we report the mean and population standard deviation (normalized by the number of runs, $n{=}5$). For the trained models the runs differ in the random seed and, on DEN, in the validation fold. The linear probe (TESSERA-LP) is fit by a deterministic solver, so on DEN and PASTIS-R its variability comes from the training subset instead: we draw $5$ independent class-preserving subsets of the training pixels (Section~\ref{sec:probe}) and fit one probe per subset, keeping the test set fixed. On LUCAS the probe is fit once on the fixed balanced training split.

\textbf{Hardware and frameworks.} The UNet models are implemented in PyTorch~2.7 (CUDA~12.6) and trained on a single NVIDIA~H100~NVL GPU ($94$\,GB); a single run takes $\approx 10$--$90$\,min depending on crop size, temporal window and whether the embeddings fit in RAM. The linear probe is instead fit on CPU with scikit-learn~1.9.0, using its \texttt{LogisticRegression} estimator with the L-BFGS deterministic solver.

\subsubsection{Compared Models}
\label{subsubsec:compared_models}
To contextualize the results of our analysis, we also
evaluate a set of convolutional baselines that do not rely on the Tessera embeddings: four UNets with the same architecture trained from scratch on per-window Sentinel composites instead of frozen embeddings, denoted UNet-RGB, UNet-RGB\textsubscript{tm}, UNet-S2 and UNet-S1S2.

Their inputs are per-window composites over the same temporal windows as the embeddings, so only the input differs from TESSERA-UNet. For Sentinel-2 we use a per-pixel temporal median, cloud-masked with the Scene Classification Layer (SCL). For Sentinel-1 (VV/VH) we use a plain temporal median, as radar is insensitive to clouds. The four variants differ in the number of input channels. UNet-RGB and UNet-RGB\textsubscript{tm} ($3$ channels) use the B04/B03/B02 bands, emulating the common practice of collapsing Sentinel-2 into an RGB image via, respectively, $z$-score normalization or a percentile tonemapping ($2$nd and $98$th percentiles, $\gamma=0.5$). UNet-S2 ($10$ channels) uses the ten Sentinel-2 bands, and UNet-S1S2 ($12$ channels) additionally includes the two Sentinel-1 channels. All inputs except UNet-RGB\textsubscript{tm} are per-channel $z$-score standardized with train-split statistics. UNet-RGB\textsubscript{tm} instead relies on the percentile tonemapping as its standardization, with percentiles computed on the train split. The full construction is detailed in Appendix~\ref{app:baseline_inputs}.

Since the median composites collapse the temporal axis within each window, we additionally train U-TAE~\cite{garnot2021panoptic}, a spatio-temporal segmentation network that attends over the acquisition dates, from scratch on the raw Sentinel-2 time series of each window. It takes the same ten Sentinel-2 bands as UNet-S2, so the two differ in whether the time series is collapsed into a composite and in the network that processes it. We use the official implementation with its default hyperparameters and the training recipe shared by all models (Section~\ref{subsubsec:training_setup}), keeping the effective batch size through gradient accumulation (micro-batches of $4$ on DEN and $16$ on PASTIS-R). The input sequence contains every Sentinel-2 acquisition in the window, without cloud masking as in the original U-TAE, and discards only dates without data. It is capped at $40$ dates, the number of observations ingested by Tessera ($n{=}40$, Section~\ref{sec:framework}): longer sequences are subsampled at random during training and uniformly during validation and test.

\subsubsection{Evaluation Metrics}
\label{subsubsec:evaluation_metrics}
All models are scored at the pixel level over the classes $\mathcal{C}$ defined by each benchmark
($|\mathcal{C}|=8$ for LUCAS, $|\mathcal{C}|=19$ for PASTIS-R and $|\mathcal{C}|=6$ for DEN), using four complementary
metrics reported as percentages: overall accuracy, balanced accuracy, macro-averaged F1
score and mean Intersection-over-Union (mIoU), the latter being the standard primary
metric for semantic segmentation.
The Appendix gives the formal definitions of the metrics and the
averaging conventions.

\subsection{Quantitative Results}
\label{subsec:quantitative_results}

\subsubsection{Main Results}
\label{subsubsec:main_results}

\begin{table*}[t]
\centering
\caption{Semantic segmentation results (\%) on the DEN and PASTIS-R (19 classes, background included) test sets across temporal windows. All models are reported as mean\,{\scriptsize$\pm$\,std} over 5 runs: random seeds for TESSERA-UNet, U-TAE and the from-scratch UNets, and 5 independently drawn training subsets for TESSERA-LP. The gray rows are trivial baselines that ignore the input (always predicting the most frequent test class, and predicting at random with the test class frequencies); they do not depend on the window. Best result per column and dataset in \fst{green}, second best in \snd{blue}.}
\label{tab:results_full_combined}
\scriptsize
\resizebox{\textwidth}{!}{%
\begin{tabular}{ll cccc cccc cccc cccc}
\toprule
& & \multicolumn{4}{c}{1M} & \multicolumn{4}{c}{3M} & \multicolumn{4}{c}{6M} & \multicolumn{4}{c}{1Y} \\
\cmidrule(lr){3-6} \cmidrule(lr){7-10} \cmidrule(lr){11-14} \cmidrule(lr){15-18}
& Method & Acc & B.Acc & F1 & mIoU & Acc & B.Acc & F1 & mIoU & Acc & B.Acc & F1 & mIoU & Acc & B.Acc & F1 & mIoU \\
\midrule
\multirow{9}{*}{\rotatebox[origin=c]{90}{DEN}}
& UNet-RGB                    & 62.9\pms{0.6} & 45.7\pms{0.7} & 43.5\pms{0.5} & 30.9\pms{0.3} & 61.5\pms{1.6} & 46.4\pms{1.4} & 44.1\pms{0.2} & 31.3\pms{0.1} & 64.1\pms{0.5} & 48.7\pms{1.1} & 46.5\pms{0.5} & 33.8\pms{0.3} & 65.9\pms{0.9} & 48.9\pms{0.9} & 47.6\pms{0.5} & 35.1\pms{0.3} \\
& UNet-RGB\textsubscript{tm}  & 63.6\pms{0.9} & 45.3\pms{1.3} & 43.9\pms{0.8} & 31.2\pms{0.6} & 63.2\pms{0.9} & 46.2\pms{1.0} & 44.7\pms{0.3} & 31.8\pms{0.3} & 64.3\pms{0.7} & 49.3\pms{1.3} & 46.8\pms{0.8} & 34.0\pms{0.6} & 66.5\pms{0.4} & 49.8\pms{0.5} & 48.4\pms{0.4} & 35.8\pms{0.2} \\
& UNet-S2                     & 72.0\pms{0.4} & \fst{58.6\pms{0.7}} & \fst{55.0\pms{0.6}} & \fst{42.4\pms{0.3}} & 71.7\pms{0.4} & \snd{57.9\pms{1.1}} & \fst{55.6\pms{0.6}} & \snd{42.6\pms{0.4}} & 72.0\pms{0.6} & \snd{58.8\pms{0.5}} & \fst{57.0\pms{0.6}} & \snd{43.9\pms{0.4}} & 71.3\pms{1.0} & \snd{58.5\pms{1.0}} & \fst{56.9\pms{0.7}} & \snd{43.5\pms{0.4}} \\
& UNet-S1S2                   & \fst{73.0\pms{0.5}} & 53.7\pms{1.2} & 52.9\pms{0.7} & \snd{41.2\pms{0.4}} & \fst{72.6\pms{0.5}} & 56.7\pms{0.7} & 54.6\pms{0.4} & 42.0\pms{0.3} & \fst{73.4\pms{0.8}} & 57.6\pms{1.1} & \snd{56.9\pms{0.8}} & \fst{44.2\pms{0.6}} & \fst{73.0\pms{0.8}} & 57.8\pms{0.8} & \snd{56.7\pms{0.6}} & \fst{43.9\pms{0.3}} \\
& U-TAE                 & \snd{72.1\pms{0.9}} & \snd{56.6\pms{1.2}} & \snd{53.5\pms{1.5}} & 41.1\pms{1.2} & \snd{72.2\pms{0.2}} & \fst{58.1\pms{1.4}} & \snd{55.0\pms{1.1}} & 42.3\pms{0.8} & 71.6\pms{0.6} & \fst{60.3\pms{1.2}} & 55.0\pms{0.9} & 42.6\pms{0.7} & 70.9\pms{0.6} & \fst{60.3\pms{1.0}} & 54.7\pms{1.0} & 41.8\pms{0.8} \\
& TESSERA-LP                  & 68.2\pms{0.1} & 50.3\pms{0.1} & 49.6\pms{0.1} & 37.9\pms{0.1} & 69.3\pms{0.0} & 52.2\pms{0.1} & 51.6\pms{0.1} & 39.3\pms{0.1} & 71.3\pms{0.0} & 53.8\pms{0.3} & 54.2\pms{0.3} & 41.4\pms{0.2} & \snd{71.6\pms{0.1}} & 54.1\pms{0.3} & 55.8\pms{0.2} & 42.0\pms{0.2} \\
& TESSERA-UNet                & 70.3\pms{0.8} & 53.6\pms{0.8} & 52.6\pms{0.4} & 40.8\pms{0.1} & 71.8\pms{0.8} & 56.4\pms{0.3} & \snd{55.0\pms{0.2}} & \fst{42.7\pms{0.2}} & \snd{72.8\pms{0.6}} & 56.5\pms{1.3} & 55.2\pms{0.9} & 43.1\pms{0.6} & 71.3\pms{0.8} & 57.3\pms{2.2} & 55.9\pms{1.1} & 42.9\pms{1.0} \\
\cmidrule(lr){2-18}
& \textcolor{gray}{Majority class} & \textcolor{gray}{48.8} & \textcolor{gray}{16.7} & \textcolor{gray}{10.9} & \textcolor{gray}{8.1} & \textcolor{gray}{48.8} & \textcolor{gray}{16.7} & \textcolor{gray}{10.9} & \textcolor{gray}{8.1} & \textcolor{gray}{48.8} & \textcolor{gray}{16.7} & \textcolor{gray}{10.9} & \textcolor{gray}{8.1} & \textcolor{gray}{48.8} & \textcolor{gray}{16.7} & \textcolor{gray}{10.9} & \textcolor{gray}{8.1} \\
& \textcolor{gray}{Random (class prior)} & \textcolor{gray}{34.8} & \textcolor{gray}{16.7} & \textcolor{gray}{16.7} & \textcolor{gray}{10.2} & \textcolor{gray}{34.8} & \textcolor{gray}{16.7} & \textcolor{gray}{16.7} & \textcolor{gray}{10.2} & \textcolor{gray}{34.8} & \textcolor{gray}{16.7} & \textcolor{gray}{16.7} & \textcolor{gray}{10.2} & \textcolor{gray}{34.8} & \textcolor{gray}{16.7} & \textcolor{gray}{16.7} & \textcolor{gray}{10.2} \\
\midrule
\multirow{9}{*}{\rotatebox[origin=c]{90}{PASTIS-R}}
& UNet-RGB                    & 63.1\pms{0.3} & 36.4\pms{0.3} & 34.1\pms{0.1} & 22.7\pms{0.1} & 64.3\pms{0.1} & 38.7\pms{0.2} & 35.5\pms{0.1} & 23.8\pms{0.1} & 64.9\pms{0.1} & 41.4\pms{0.3} & 38.1\pms{0.1} & 25.7\pms{0.1} & 63.9\pms{0.2} & 41.3\pms{0.4} & 37.7\pms{0.2} & 25.3\pms{0.2} \\
& UNet-RGB\textsubscript{tm}  & 63.1\pms{0.6} & 36.5\pms{0.2} & 34.4\pms{0.2} & 22.9\pms{0.1} & 64.2\pms{0.1} & 38.5\pms{0.1} & 35.4\pms{0.1} & 23.7\pms{0.0} & 64.8\pms{0.1} & 41.0\pms{0.1} & 37.8\pms{0.0} & 25.5\pms{0.0} & 63.9\pms{0.1} & 41.1\pms{0.2} & 37.5\pms{0.2} & 25.2\pms{0.1} \\
& UNet-S2                     & 69.5\pms{0.3} & 48.9\pms{0.2} & 45.7\pms{0.4} & 32.0\pms{0.3} & 69.6\pms{0.3} & 49.0\pms{0.4} & 45.5\pms{0.2} & 31.8\pms{0.1} & 70.4\pms{0.4} & 51.3\pms{0.4} & 47.5\pms{0.2} & 33.6\pms{0.2} & 72.1\pms{0.6} & 55.2\pms{0.4} & 51.3\pms{0.1} & 37.5\pms{0.1} \\
& UNet-S1S2                   & \snd{70.7\pms{0.2}} & \snd{50.5\pms{0.2}} & \snd{47.1\pms{0.2}} & \snd{33.3\pms{0.2}} & \snd{71.1\pms{0.1}} & 50.7\pms{0.4} & 47.4\pms{0.2} & 33.5\pms{0.2} & \snd{72.0\pms{0.1}} & 52.9\pms{0.3} & 49.8\pms{0.1} & 35.7\pms{0.1} & 74.3\pms{0.1} & 57.2\pms{0.4} & 53.8\pms{0.3} & 40.0\pms{0.2} \\
& U-TAE                 & 63.5\pms{1.4} & 48.4\pms{2.5} & 42.7\pms{2.0} & 29.3\pms{1.6} & 67.8\pms{0.6} & \fst{58.9\pms{0.5}} & \snd{50.2\pms{0.4}} & \snd{35.9\pms{0.4}} & 69.9\pms{0.9} & \fst{64.1\pms{0.6}} & \snd{54.7\pms{0.7}} & \snd{40.0\pms{0.6}} & 74.2\pms{0.8} & \fst{77.3\pms{0.7}} & \snd{65.7\pms{0.5}} & \snd{51.4\pms{0.5}} \\
& TESSERA-LP                  & 58.7\pms{0.1} & 29.2\pms{0.1} & 27.5\pms{0.1} & 18.0\pms{0.1} & 65.1\pms{0.0} & 41.8\pms{0.2} & 39.3\pms{0.1} & 26.7\pms{0.0} & 70.4\pms{0.1} & 53.0\pms{0.2} & 49.0\pms{0.2} & 35.2\pms{0.1} & \snd{76.1\pms{0.0}} & 67.2\pms{0.1} & 62.6\pms{0.2} & 48.4\pms{0.2} \\
& TESSERA-UNet                & \fst{72.1\pms{0.4}} & \fst{52.5\pms{1.0}} & \fst{49.5\pms{0.2}} & \fst{35.5\pms{0.2}} & \fst{74.8\pms{0.1}} & \snd{58.3\pms{0.4}} & \fst{55.7\pms{0.1}} & \fst{41.0\pms{0.1}} & \fst{76.6\pms{0.1}} & \snd{62.1\pms{0.5}} & \fst{60.0\pms{0.3}} & \fst{45.3\pms{0.3}} & \fst{80.6\pms{0.1}} & \snd{74.0\pms{0.4}} & \fst{71.9\pms{0.3}} & \fst{58.3\pms{0.4}} \\
\cmidrule(lr){2-18}
& \textcolor{gray}{Majority class} & \textcolor{gray}{44.1} & \textcolor{gray}{5.3} & \textcolor{gray}{3.2} & \textcolor{gray}{2.3} & \textcolor{gray}{44.1} & \textcolor{gray}{5.3} & \textcolor{gray}{3.2} & \textcolor{gray}{2.3} & \textcolor{gray}{44.1} & \textcolor{gray}{5.3} & \textcolor{gray}{3.2} & \textcolor{gray}{2.3} & \textcolor{gray}{44.1} & \textcolor{gray}{5.3} & \textcolor{gray}{3.2} & \textcolor{gray}{2.3} \\
& \textcolor{gray}{Random (class prior)} & \textcolor{gray}{25.3} & \textcolor{gray}{5.3} & \textcolor{gray}{5.3} & \textcolor{gray}{3.0} & \textcolor{gray}{25.3} & \textcolor{gray}{5.3} & \textcolor{gray}{5.3} & \textcolor{gray}{3.0} & \textcolor{gray}{25.3} & \textcolor{gray}{5.3} & \textcolor{gray}{5.3} & \textcolor{gray}{3.0} & \textcolor{gray}{25.3} & \textcolor{gray}{5.3} & \textcolor{gray}{5.3} & \textcolor{gray}{3.0} \\
\bottomrule
\end{tabular}%
}
\end{table*}

Table~\ref{tab:results_full_combined} gathers the results of all the trained models on the DEN and PASTIS-R datasets, reporting semantic segmentation metrics over 1 month (1M), 3 months (3M), 6 months (6M) and 1 year (1Y) temporal windows.

Across the three benchmarks, Tessera embeddings preserve much of their value as the observation window contracts, but how much they retain, and how much they help in the first place, varies between tasks in a way that is consistent with how strongly their classes rely on phenology rather than on spectral appearance, although the benchmarks also differ in other respects (Section~\ref{sec:discussion}). We first quantify this at the full-year window and then trace how performance degrades as the window shrinks, down to a single day.

\paragraph{The Value of the Embeddings Is Task-Dependent}
The two benchmarks rely on different signals. PASTIS-R requires distinguishing $18$ crop types that stay spectrally similar for much of the year and are separated mainly by their \emph{phenology}, a signal that lives along the temporal axis. The median composites fed to the from-scratch UNets collapse that axis, whereas the Tessera embeddings and U-TAE both keep the observation series. A temporal baseline trained from scratch already closes a large part of the gap: at 1Y, U-TAE trained from scratch on the Sentinel-2 series reaches $51.4$ mIoU, against $37.5$ for UNet-S2 on the same bands, a gain that reflects both access to the series and the different network that processes it.TESSERA-UNet adds a further $6.9$ points ($58.3$ mIoU, $13\%$ above U-TAE and $46\%$ above the best composite-based UNet) and has the highest mIoU at every window. This is an advantage over a temporal baseline trained from scratch under the same protocol. Since U-TAE uses Sentinel-2 only and a different architecture, it does not isolate the contribution of pretraining. The Sentinel-1 input, which U-TAE does not use, is worth $2.5$ points on composites (UNet-S2 vs.\ UNet-S1S2). U-TAE, in turn, has the highest balanced accuracy from 3M to 1Y.

DEN, in contrast, comprises six coarse, temporally stable land-cover classes largely separable from spectral cues, so a single composite already carries most of the signal. At full supervision the embeddings bring no gain over raw S2/S1+S2 composites (within $2.5\%$ of the best from-scratch UNet, $43.1$ vs.\ $44.2$ mIoU at 6M), and the RGB variants trail only for lack of spectral bands (${\sim}35$ mIoU at 1Y).

The embeddings thus cost little when they do not help and win large when they do. Even on DEN they regain a clear edge once labels are scarce (Section~\ref{subsubsec:label_efficiency}).

\paragraph{Sensitivity to the Temporal Window}
Shrinking the observation window exposes the same split: TESSERA-UNet drops by $39\%$ on PASTIS-R ($58.3\rightarrow35.5$ mIoU, 1Y$\rightarrow$1M) but by only $5\%$ on DEN ($42.9\rightarrow40.8$). Although these percentages include results from samples that might extend the acquisition window, we show that on DEN, the difference between scoped and unscoped sets is $3.4\%$ and $10.5\%$ respectively (see Appendix~\ref{app:observations}). That embeddings from a model trained on annual series remain this informative over much shorter windows is notable, and unlocks operating regimes with limited temporal coverage such as low-latency mapping and faster LULC refresh cycles.

U-TAE degrades faster: from 1Y to 1M it loses $43\%$ of its mIoU on PASTIS-R ($51.4\rightarrow29.3$) and falls below UNet-S1S2 ($33.3$), whereas TESSERA-UNet keeps $35.5$. With only a few dates per window, a temporal model trained from scratch has little sequence to learn from, whereas the frozen embeddings come from an encoder pretrained on full-year series, which is consistent with the relative advantage over U-TAE being largest at the shortest window ($21\%$ at 1M vs.\ $13\%$ at 1Y). On DEN, U-TAE stays between $41.1$ and $42.6$ mIoU at every window, in the range of the composite-based UNets and TESSERA-UNet.

\paragraph{A Linear Probe Suffices When Classes Are Linearly Separable}
On DEN the linear probe nearly matches the UNet head ($41.4$ vs.\ $43.1$ mIoU at 6M; $42.0$ vs.\ $42.9$ at 1Y), so at this coarse taxonomy the classes are almost linearly separable in embedding space, enabling lightweight decoders. On PASTIS-R the probe is competitive only with enough temporal context: at 1Y it reaches $48.4$ mIoU, beating every composite-based UNet by $21\%$, but it collapses at short windows, where the spatial aggregation of the head becomes essential.

\begin{figure*}[t]
    \centering
    \includegraphics[width=1\linewidth]{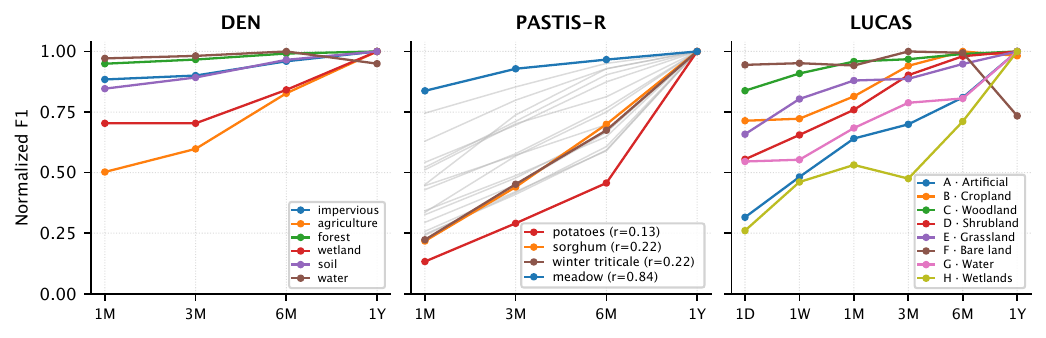}
    \caption{Per-class temporal sensitivity of the linear probe (TESSERA-LP, fit on a single training subset): per-class F1 normalized by its maximum across windows. DEN and LUCAS show all classes; for PASTIS-R (19 classes) we highlight the most and least temporally sensitive classes, ranked by the F1 retained at the shortest window relative to 1Y ($r=\mathrm{F1(1M)}/\mathrm{F1(1Y)}$), and show the remaining classes in gray.}
    \label{fig:per_class}
\end{figure*}

\paragraph{Per-Class Temporal Sensitivity}
Measured with the linear probe (TESSERA-LP), the dataset-level dichotomy also holds \emph{within} each benchmark at the class level (Figure~\ref{fig:per_class}). Phenology-driven classes degrade steeply as the window shrinks, whereas structurally or spectrally distinct classes stay almost flat.

On DEN, agriculture loses about half of its 1Y F1 by 1M, while forest and water are essentially window-invariant. The split is sharpest on PASTIS-R, where the most temporally sensitive classes (crops such as potatoes, sorghum and winter triticale) retain barely $13$--$22\%$ of their 1Y F1 at 1M, whereas the comparatively stable meadow retains $84\%$. LUCAS confirms the trend down to a single day: Cropland and Grassland degrade markedly faster than the structurally distinct Woodland. Bare land runs against the trend, slightly improving at shorter windows, as its largely time-invariant spectral signature is diluted by temporal compositing.

A few classes (DEN wetland, LUCAS water and wetlands) stay low at every window; their low scores are more plausibly explained by severe under-representation and small, sparse polygons near Sentinel's $10$\,m resolution than by temporal coverage, so longer temporal coverage alone cannot resolve them. Overall, temporal requirements are class-dependent: the classes that degrade most are those with a marked seasonal cycle. Since these comparisons are made within a single benchmark, they hold geography, observation conditions and annotation protocol fixed, and are less confounded than the comparison between benchmarks. This pattern motivates class-adaptive temporal windows in practice.

\paragraph{Temporal Resolution on LUCAS}
Unlike DEN and PASTIS-R, LUCAS provides sparse polygons rather than dense masks, making it a pixel-wise \emph{classification} task. Training a UNet is not meaningful, so we evaluate only the linear probe (Section~\ref{sec:method}), which lets us push the temporal resolution down to a single day.

Performance decreases monotonically as the window shrinks but far less than the loss of context would suggest (Table~\ref{tab:temporal_resolution}). In balanced accuracy, halving from 1Y ($50.8\%$) to 6M ($49.8\%$) is negligible and 1M stays reasonable ($44.6\%$). Only under extreme sparsity does it drop markedly, to $36.8\%$ at a 1-day window ($27.6\%$ below the yearly baseline), yet still $2.9$ times the $12.5\%$ chance level of balanced accuracy for the eight classes. Overall accuracy is less informative here: because the test set is imbalanced, always predicting cropland already reaches $34.5\%$, so the $43.0\%$ at 1D is $8.5$ points above this trivial baseline. The LUCAS test set varies slightly with the window: a polygon has no embedding when no Sentinel-2 acquisition in the window covers it with valid data, which leaves $13{,}565$ of the $14{,}200$ test polygons at 1D and all of them at 6M and 1Y (and $757$ to $800$ training polygons). Evaluating every window on the $13{,}560$ polygons common to all of them changes no metric of Table~\ref{tab:temporal_resolution} by more than $0.46$ points, and the drop in overall accuracy from 1Y to 1D by $0.01$ points.

Single-date embeddings thus retain meaningful land-cover information, confirming on a third, pan-European dataset that Tessera representations preserve much of their semantics under severe temporal sparsity~\cite{feng2025tessera}.

\begin{table}[tb]
\centering
\caption{Classification performance (\%) in LUCAS dataset as a function of the temporal
resolution of the input. Best value per column in \fst{green}, second best in
\snd{blue}. F1$_\mathrm{w}$ denotes the support-weighted F1 score; all other
class averages are macro-averaged. The two bottom rows are trivial baselines that ignore the input: always predicting the most frequent test class (cropland, $34.5\%$ of the test polygons), and predicting at random with the class frequencies of the test set (expected values); they are computed on the 1Y test pixels.}
\label{tab:temporal_resolution}
\small
\setlength{\tabcolsep}{6pt}
\begin{tabular}{l ccccc}
\toprule
Temporal resolution & Acc. & Bal. Acc. & Prec. & F1 & F1$_\mathrm{w}$ \\
\midrule
1 Day    & 43.0 & 36.8 & 29.2 & 25.8 & 49.2 \\
1 Week   & 47.5 & 42.1 & 31.8 & 28.8 & 53.5 \\
1 Month  & 52.3 & 44.6 & 33.9 & 31.7 & 58.3 \\
3 Months & 55.5 & 46.4 & 35.6 & 33.9 & 61.8 \\
6 Months & \snd{58.8} & \snd{49.8} & \snd{37.3} & \snd{35.8} & \snd{64.8} \\
1 Year   & \fst{59.5} & \fst{50.8} & \fst{38.1} & \fst{36.5} & \fst{65.4} \\
\midrule
Majority class & 34.5 & 12.5 & 4.3 & 6.4 & 17.7 \\
Random (class prior) & 28.6 & 12.5 & 12.5 & 12.5 & 28.6 \\
\bottomrule
\end{tabular}
\end{table}

\subsubsection{Published PASTIS-R Results for Context}
\label{subsubsec:published_context}
Table~\ref{tab:sota_pastis} places our models alongside the results published on PASTIS-R at full labels~\cite{feng2025tessera}. It provides context rather than a head-to-head comparison. The foundation-model baselines were evaluated in~\cite{feng2025tessera} with PANGAEA~\cite{marsocci2024pangaea}, under its own data preparation, temporal sampling and decoders, and the downstream protocol behind the published Tessera and AlphaEarth results is only partly described, so we could not reproduce it. Several settings may therefore differ from ours, and each of them affects the reported mIoU: (i)~the data split, where we use the official folds (Section~\ref{sec:datasets}) while the split behind the published Tessera and AlphaEarth results is not stated; (ii)~the scored classes, where we score the background and the $18$ crop types; (iii)~the aggregation of the metric, where we accumulate a single confusion matrix over all test pixels (Appendix~\ref{app:metrics}), whereas averaging the mIoU over test patches instead gives $41.3$ rather than $58.3$ for TESSERA-UNet on the same predictions; (iv)~the downstream head and training recipe; and (v)~the Sentinel-1/2 inputs and observation windows, which we build from our own acquisitions (Section~\ref{sec:framework}). Within Table~\ref{tab:sota_pastis}, only U-TAE, TESSERA-LP and TESSERA-UNet share the data, split, training recipe and metric, and their comparison is discussed in Section~\ref{subsubsec:main_results}. We draw no conclusion from the published rows about the relative performance of the models.

We report no analogous comparison for DEN. It is originally a change-detection benchmark, and we evaluate it as a $6$-class land-cover segmentation task with labels resampled from $3$\,m to $10$\,m, a setup for which no directly comparable published numbers exist. We therefore rely on the from-scratch UNet baselines of Table~\ref{tab:results_full_combined} as the reference point on DEN.

\begin{table}[tb]
\centering
\caption{Semantic segmentation on PASTIS-R (mIoU, full labels). Foundation-model baselines are reported both original and finetuned, as in~\cite{feng2025tessera}; AlphaEarth, Tessera and our TESSERA-based models use frozen embeddings with a trained head/probe, and U-TAE~\cite{garnot2021panoptic} is trained from scratch on the 1Y window. Ours are mean\,{\scriptsize$\pm$\,std} over 5 runs. Reported for context: only our rows share a common protocol (Section~\ref{subsubsec:published_context}).}
\label{tab:sota_pastis}
\small
\setlength{\tabcolsep}{8pt}
\begin{tabular}{l cc}
\toprule
Model & Original & Finetuned \\
\midrule
CROMA~\cite{fuller2023croma}        & 36.24 & 41.29 \\
DOFA~\cite{xiong2024DOFA} & 25.94 & 29.55 \\
Prithvi~\cite{szwarcman2025prithvi} & 34.09 & 39.52 \\
RemoteCLIP~\cite{remoteclip}        & 20.17 & 21.03 \\
SatlasNet~\cite{Bastani2023}        & 21.02 & 31.38 \\
Scale-MAE~\cite{Reed2023}           & 23.96 & 28.51 \\
SpectralGPT~\cite{SpectralGPT}      & 35.10 & 40.28 \\
Galileo~\cite{tseng2025galileolearningglobal} & 27.92 & 35.28 \\
Skysense~\cite{Guo2024}             & 32.87 & 36.41 \\
UNet Baseline~\cite{ronnebergerUNetConvolutionalNetworks2015} & 30.16 & 37.56 \\
ViT Baseline~\cite{dosovitskiy2020image} & 37.37 & 42.57 \\
\midrule
AlphaEarth + head~\cite{brown2025alphaearth}   & \multicolumn{2}{c}{51.08} \\
Tessera + head~\cite{feng2025tessera} & \multicolumn{2}{c}{50.68} \\
\midrule
\textbf{U-TAE (ours, from scratch)} & \multicolumn{2}{c}{51.4\pms{0.5}} \\
\textbf{TESSERA-LP (ours)} & \multicolumn{2}{c}{48.4\pms{0.2}} \\
\textbf{TESSERA-UNet (ours)} & \multicolumn{2}{c}{58.3\pms{0.4}} \\
\bottomrule
\end{tabular}
\end{table}

\subsubsection{Label Efficiency}
\label{subsubsec:label_efficiency}

\begin{figure*}[t]
    \centering
    \includegraphics[width=1\linewidth]{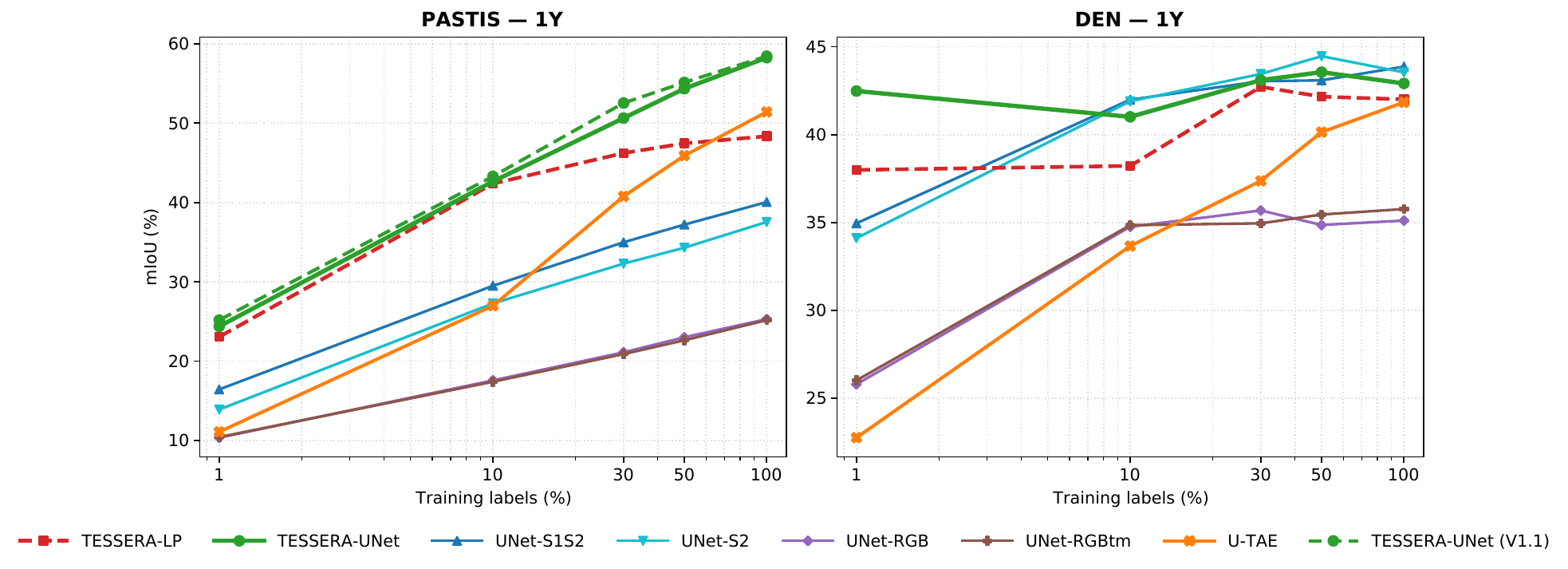}
    \caption{Label efficiency at the 1Y window. Test mIoU (\%) as a function of the fraction of training labels ($1$--$100\%$) on PASTIS-R (left) and DEN (right). On PASTIS-R we additionally overlay the V1.1 embeddings decoded with TESSERA-UNet (dashed). V1.1 is evaluated on PASTIS-R only. U-TAE is trained from scratch on the raw Sentinel-2 series. For every model, points below $100\%$ are single runs (one seed, or one training subset for TESSERA-LP); points at $100\%$ are the means over 5 runs of Table~\ref{tab:results_full_combined}.}
    \label{fig:label_efficiency}
\end{figure*}

The results reported in Section~\ref{subsubsec:main_results} use the full training set. Since frozen foundation-model embeddings are expected to be most valuable under limited supervision, we also evaluate all models while varying the fraction of training labels from $1\%$ to $100\%$. We fix the observation window at 1Y, the best-performing temporal configuration in Table~\ref{tab:results_full_combined}. Figure~\ref{fig:label_efficiency} reports test mIoU as a function of the label budget on both benchmarks. The budget is counted in annotated patches on PASTIS-R ($13$, $142$, $432$ and $723$ of the $1{,}452$ training patches for $1$, $10$, $30$ and $50\%$) and in annotated tile-months on DEN ($10$, $116$, $344$ and $571$ of $1{,}143$); the fraction of labeled pixels matches the nominal fraction within $0.3$ points. The subsets are drawn once, are nested ($1\%\subset10\%\subset30\%\subset50\%$) and are shared by all models, and the linear probe is fit on them with the same protocol as at full supervision. They are not stratified by class, so at $1\%$ two crop classes of PASTIS-R (potatoes and orchard) have no training samples; from $10\%$ on, the class proportions are close to those of the full training set.

On PASTIS-R, the TESSERA-based models lead at every label budget. With only $1\%$ of the labels, TESSERA-UNet already reaches ${\sim}24$ mIoU, roughly $50\%$ higher than the best composite-based UNet (UNet-S1S2) and more than twice U-TAE ($11.1$), which has too few labeled sequences to learn temporal features from scratch.  As in Fig.~\ref{fig:label_efficiency}, these points below $100\%$ are single runs, and at $1\%$ two PASTIS-R classes (potatoes and orchard) have no training samples, so this comparison is less certain than those at full supervision. U-TAE closes most of the gap as labels grow ($40.8$ mIoU at $30\%$, $51.4$ at $100\%$), so the advantage over this temporal baseline trained from scratch is largest when labels are scarce. TESSERA-LP outperforms every composite-based UNet at every label budget, and with $10\%$ of the labels it already exceeds the best composite-based UNet trained on all labels ($42.5$ vs.\ $40.0$ mIoU); it also stays ahead of U-TAE up to $50\%$ of the labels. The frozen embeddings are thus drastically more label-efficient than training from scratch, whether from composites or from the time series, consistent with the temporal signal that separates the crop types being already encoded in the representation.

DEN reveals the complementary picture. Here the TESSERA models start high at $1\%$ (${\sim}38$--$42$ mIoU) and stay roughly flat, whereas the composite-based UNets start much lower (${\sim}26$--$35$) and only catch up around $30$--$100\%$ of the labels, and U-TAE starts lowest ($22.8$) and approaches the other models only with all labels. The apparent parity on DEN at full labels (Table~\ref{tab:results_full_combined}) is therefore a parity at saturation. At low label budgets the frozen embeddings retain a clear advantage even for this spectrally-driven, temporally-stable task. In short, on the benchmarks evaluated here, TESSERA embeddings help when the classes depend on phenology or labels are scarce, and lose their edge only when the classes are spectrally distinct and labels are abundant.

Finally, Figure~\ref{fig:label_efficiency} also overlays the updated Tessera V1.1 embeddings on PASTIS-R (dashed). The two versions track each other across the whole curve, so at this window the update brings no label-efficiency advantage. We examine the V1.1 release in detail, including shorter windows at full supervision, in Section~\ref{subsubsec:v11_analysis}.

\subsubsection{Ablation Studies}
\label{subsubsec:ablation_studies}
TESSERA-UNet applies, to the best of our knowledge, the same model as the original Tessera work~\cite{feng2025tessera}, i.e., its frozen embeddings decoded with the UNet head defined by the authors, to datasets beyond those originally evaluated. Since not all implementation details were public, some design decisions had to be inferred. This section presents ablation studies to verify that those decisions indeed improve performance. Table~\ref{tab:ablation_DEN} and Table~\ref{tab:ablation_PASTIS} report the incremental impact of input normalization, \emph{inverse-sqrt} class weighting, and extended training (from $30$ to $80$ epochs) on DEN 6M and PASTIS-R 1Y, respectively. The best configuration, selected by validation mIoU (Section~\ref{subsubsec:training_setup}), is the one used for all TESSERA-UNet results reported in Table~\ref{tab:results_full_combined}; the tables report the corresponding test metrics.

Two observations emerge from Tables~\ref{tab:ablation_DEN} and~\ref{tab:ablation_PASTIS}. First, class weighting trades overall accuracy for class-balanced performance. On DEN it costs under $1\%$ in overall accuracy but improves balanced accuracy by $14\%$ and mIoU by $9\%$. On PASTIS-R it initially degrades mIoU (by ${\sim}1.5\%$), which is only recovered by the longer training schedule. Second, our baseline configuration already reaches $57.2$ mIoU on PASTIS-R, so the additional ingredients bring only a marginal further gain there ($57.2 \rightarrow 58.3$, $+1.9\%$). On DEN, not covered by the original recipe, they instead yield an $8.8\%$ gain ($39.6 \rightarrow 43.1$), driven mostly by the class weighting.

\begin{table}[tb]
    \centering
    \small
    \setlength{\tabcolsep}{4pt}
    \caption{Ablation study on DEN 6M for \textbf{TESSERA-UNet},
    regarding the incremental impact of input normalization,
    \emph{inverse-sqrt} class weighting, and extended training (80 epochs) on
    segmentation metrics (\%). The best configuration, selected by validation mIoU, is
    shown in bold; test metrics are reported. The first three rows are single runs (seed 0) trained for $30$ epochs; the last row is the configuration used in Table~\ref{tab:results_full_combined} (mean over 5 seeds, $80$ epochs).}
    \label{tab:ablation_DEN}
    \begin{tabular}{lcccc}
    \toprule
    Config & Acc. & Bal. Acc. & F1-macro & mIoU \\
    \midrule
    baseline                & 74.1 & 48.9 & 50.1 & 39.6 \\
    + normalization         & 73.7 & 48.8 & 49.6 & 39.2 \\
    + inverse\_sqrt         & 73.2 & 55.7 & 54.4 & 42.8 \\
    + 80 epochs (best)      & \textbf{72.8} & \textbf{56.5} & \textbf{55.2} & \textbf{43.1} \\
    \bottomrule
    \end{tabular}
\end{table}
\begin{table}[tb]
    \centering
    \small
    \setlength{\tabcolsep}{4pt}
    \caption{Ablation study on PASTIS-R 1Y for \textbf{TESSERA-UNet},
    regarding the incremental impact of input normalization,
    \emph{inverse-sqrt} class weighting, and extended training (80 epochs) on
    segmentation metrics (\%). The best configuration, selected by validation mIoU, is
    shown in bold; test metrics are reported. The first three rows are single runs (seed 0) trained for $30$ epochs; the last row is the configuration used in Table~\ref{tab:results_full_combined} (mean over 5 seeds, $80$ epochs).}
    \label{tab:ablation_PASTIS}
    \begin{tabular}{lcccc}
    \toprule
    Config & Acc. & Bal. Acc. & F1-macro & mIoU \\
    \midrule
    baseline                & 81.5 & 69.5 & 70.7 & 57.2 \\
    + normalization         & 81.6 & 69.9 & 71.1 & 57.7 \\
    + inverse\_sqrt         & 79.9 & 76.3 & 70.7 & 56.8 \\
    + 80 epochs (best)      & \textbf{80.6} & \textbf{74.0} & \textbf{71.9} & \textbf{58.3} \\
    \bottomrule
    \end{tabular}
\end{table}

\subsubsection{Impact of the Tessera V1.1 Update}
\label{subsubsec:v11_analysis}

\begin{table}[!t]
\centering
\caption{TESSERA V1 vs.\ V1.1 embeddings on PASTIS-R across temporal windows (test \%), decoded with the TESSERA-UNet head (mean\,{\scriptsize$\pm$\,std} over 5 seeds) and with a linear probe (TESSERA-LP, mean\,{\scriptsize$\pm$\,std} over 5 training-subset draws). The larger value per version pair and metric is in bold.}
\label{tab:v11_vs_v1}
\scriptsize
\setlength{\tabcolsep}{4pt}
\resizebox{\columnwidth}{!}{%
\begin{tabular}{ll cccc}
\toprule
Window & Embeddings & Acc & B.Acc & F1 & mIoU \\
\midrule
\multirow{4}{*}{1M}
& TESSERA-UNet (V1)   & \textbf{72.1}\pms{0.4} & \textbf{52.5}\pms{1.0} & \textbf{49.5}\pms{0.2} & \textbf{35.5}\pms{0.2} \\
& TESSERA-UNet (V1.1) & 71.9\pms{0.2} & 51.8\pms{0.7} & 49.1\pms{0.2} & 35.1\pms{0.1} \\
& TESSERA-LP (V1)     & 58.7\pms{0.1} & \textbf{29.2}\pms{0.1} & \textbf{27.5}\pms{0.1} & 18.0\pms{0.1} \\
& TESSERA-LP (V1.1)   & \textbf{58.9}\pms{0.1} & 29.1\pms{0.3} & \textbf{27.5}\pms{0.2} & \textbf{18.1}\pms{0.1} \\
\midrule
\multirow{4}{*}{3M}
& TESSERA-UNet (V1)   & \textbf{74.8}\pms{0.1} & \textbf{58.3}\pms{0.4} & \textbf{55.7}\pms{0.1} & \textbf{41.0}\pms{0.1} \\
& TESSERA-UNet (V1.1) & 74.4\pms{0.1} & 57.6\pms{0.3} & 54.8\pms{0.3} & 40.3\pms{0.2} \\
& TESSERA-LP (V1)     & \textbf{65.1}\pms{0.0} & 41.8\pms{0.2} & \textbf{39.3}\pms{0.1} & \textbf{26.7}\pms{0.0} \\
& TESSERA-LP (V1.1)   & 64.9\pms{0.0} & \textbf{41.9}\pms{0.2} & 39.1\pms{0.1} & 26.6\pms{0.1} \\
\midrule
\multirow{4}{*}{6M}
& TESSERA-UNet (V1)   & \textbf{76.6}\pms{0.1} & 62.1\pms{0.5} & 60.0\pms{0.3} & \textbf{45.3}\pms{0.3} \\
& TESSERA-UNet (V1.1) & 76.5\pms{0.0} & \textbf{62.4}\pms{0.3} & \textbf{60.1}\pms{0.1} & \textbf{45.3}\pms{0.1} \\
& TESSERA-LP (V1)     & \textbf{70.4}\pms{0.1} & \textbf{53.0}\pms{0.2} & \textbf{49.0}\pms{0.2} & \textbf{35.2}\pms{0.1} \\
& TESSERA-LP (V1.1)   & \textbf{70.4}\pms{0.0} & 52.9\pms{0.3} & \textbf{49.0}\pms{0.2} & 35.1\pms{0.2} \\
\midrule
\multirow{4}{*}{1Y}
& TESSERA-UNet (V1)   & \textbf{80.6}\pms{0.1} & \textbf{74.0}\pms{0.4} & 71.9\pms{0.3} & 58.3\pms{0.4} \\
& TESSERA-UNet (V1.1) & \textbf{80.6}\pms{0.1} & 73.7\pms{0.4} & \textbf{72.0}\pms{0.4} & \textbf{58.4}\pms{0.3} \\
& TESSERA-LP (V1)     & 76.1\pms{0.0} & 67.2\pms{0.1} & 62.6\pms{0.2} & 48.4\pms{0.2} \\
& TESSERA-LP (V1.1)   & \textbf{76.3}\pms{0.1} & \textbf{67.8}\pms{0.2} & \textbf{63.0}\pms{0.2} & \textbf{48.7}\pms{0.2} \\
\bottomrule
\end{tabular}%
}
\end{table}

During the development of this work, an updated version of the Tessera embeddings (V1.1) was released. V1.1 widens the encoder and adds an MLP dimensionality reducer, replaces V1's random sampling of $40$ timesteps per pixel with inference over every valid observation, and normalizes each Sentinel-1 orbit direction separately\footnote{See the Tessera repository for the full V1.1 changelog: \url{https://github.com/ucam-eo/tessera}.}. Since all experiments in this paper use V1 embeddings, we verified whether this update would alter our conclusions on the benchmark where the embeddings are most decisive, PASTIS-R, at every observation window.

\paragraph{Downstream Parity}
As Table~\ref{tab:v11_vs_v1} shows, V1.1 performs comparably to V1 for the TESSERA-UNet head at every window. At 6M and 1Y the two versions differ by at most $0.3$ points on any metric, a margin comparable to the seed-to-seed standard deviation of either version. At the short windows V1 is marginally ahead, by $0.4$ mIoU at 1M and $0.7$ at 3M.
The same holds for the linear probe, where V1 and V1.1 differ by at most $0.3$ mIoU at any window, confirming that the update does not enrich the embeddings even when read out linearly, i.e., without a head that could absorb the difference. This parity is, moreover, not an artefact of operating near the task ceiling at full supervision: across the entire label-efficiency curve at the 1Y window (Figure~\ref{fig:label_efficiency}, dashed), V1.1 tracks V1 down to $1\%$ of the labels, the low-supervision regime in which a richer representation should matter most.

This parity is consistent with the Tessera authors' own positioning of the update, which they present not as a generally stronger representation but as a fix for a tiling artefact that V1 can introduce when embedding full tiles, an issue orthogonal to the per-pixel downstream tasks studied here. We therefore report V1 throughout and expect our findings to transfer unchanged to V1.1.

\paragraph{Linear Class Separability}
As an additional, decoder-agnostic check, Table~\ref{tab:lda_separability} reports the linear (LDA) class separability of the frozen embeddings, defined as $\mathrm{Tr}(\sigma_b)/\mathrm{Tr}(\sigma_w)$, the ratio of between- to within-class scatter (higher is better). Consistent with the on-par performance above, V1.1 does not meaningfully enrich the linearly-accessible class structure at the windows that matter: separability is essentially unchanged at 6M and 1Y ($-8.8\%$ and $-1.9\%$). It does rise in relative terms at 1M and 3M ($+33.8\%$ and $+43.6\%$), but in absolute terms separability at these short windows stays far below the 1Y level regardless of version (e.g., $0.048$ at 1M vs.\ $0.278$ at 1Y), and Table~\ref{tab:v11_vs_v1} shows that the relative gain does not translate into accuracy: at 1M and 3M neither head improves with V1.1 beyond the run-to-run variability. This corroborates, independently of the decoder, why the update yields no measurable benefit.

\begin{table}[tb]
\centering
\caption{Linear (LDA) class separability of the frozen embeddings on PASTIS-R (higher is better). $\Delta$ and $\Delta\%$ are the V1.1\,$-$\,V1 change; gains in \textcolor{gainpos}{green}, losses in \textcolor{lossneg}{red}.}
\label{tab:lda_separability}
\small
\setlength{\tabcolsep}{8pt}
\begin{tabular}{l c c r r}
\toprule
Window & V1 & V1.1 & $\Delta$ & $\Delta\%$ \\
\midrule
1M & 0.036 & \textbf{0.048} & \dpos{0.012} & \dpos{33.8\%} \\
3M & 0.052 & \textbf{0.075} & \dpos{0.023} & \dpos{43.6\%} \\
6M & \textbf{0.116} & 0.106 & \dneg{0.010} & \dneg{8.8\%} \\
1Y & \textbf{0.284} & 0.278 & \dneg{0.005} & \dneg{1.9\%} \\
\bottomrule
\end{tabular}
\end{table}

\subsection{Disentangling Span, Season and Number of Observations}
\label{subsubsec:temporal_factors}
Shortening the window reduces at once the temporal span, the part of the season that is observed and the number of observations (Section~\ref{sec:framework}). On PASTIS-R, where the window has the largest effect, we separate these factors with three additional analyses with the models and protocol of Table~\ref{tab:results_full_combined}. The full results are given in Appendix~\ref{app:temporal_factors}.

\begin{figure*}[t]
    \centering
    \includegraphics[width=\linewidth]{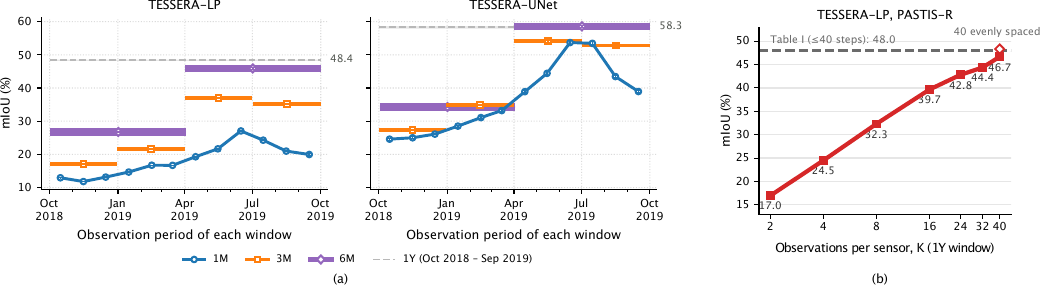}
    \caption{Factors behind the temporal sensitivity on PASTIS-R. (a)~Test mIoU of TESSERA-LP and TESSERA-UNet for each window position; each horizontal segment spans the observation period of a window, with the marker at its center, and the dashed line is the 1Y result. (b)~Test mIoU of TESSERA-LP at the 1Y window as a function of the number $K$ of observations per sensor, drawn at random within the window, on the $360$ test windows that admit $K{=}40$; the open marker uses $40$ evenly spaced dates, as in Table~\ref{tab:results_full_combined}, and the dashed line is the Table~\ref{tab:results_full_combined} probe on the same windows. Error bars ($\pm1$ std over 5 runs) are mostly smaller than the markers.}
    \label{fig:temporal_factors}
\end{figure*}

\paragraph{Seasonal Coverage}
Table~\ref{tab:results_full_combined} averages each window length over all its positions in the year. Scoring each position separately (Fig.~\ref{fig:temporal_factors}a) shows that when the window is placed matters as much as how long it is. With TESSERA-UNet, a single month reaches $53.7$ mIoU in June, $92\%$ of the 1Y value, against $24.6$ in October; the three months from April to June reach $54.1$, and the six months from April to September reach $58.6$, comparable to the full year ($58.3$). The linear probe follows the same pattern at a lower level ($27.1$ in June and $45.9$ from April to September, against $48.4$ at 1Y). The cost of a 1M window reported in Table~\ref{tab:results_full_combined} ($-39\%$) is therefore an average over the year: it is $-8\%$ if the month is June and $-58\%$ if it is October. The effect is concentrated in crops with a marked seasonal cycle (Fig.~\ref{fig:window_position_classes}). With TESSERA-UNet, potatoes are not identified by any window that ends before April (IoU below $9$, against $50.5$ at 1Y), whereas meadow reaches at least $54.6$ at every window ($62.5$ at 1Y).

\begin{table}[t]
\centering
\caption{Test mIoU of TESSERA-LP on PASTIS-R with all available observations, as in Table~\ref{tab:results_full_combined}, and with a fixed number $K{=}4$ of observations per sensor at every window. Both rows are evaluated on the test windows that admit $K{=}4$ ($3{,}425$ of $5{,}952$ at 1M, $1{,}877$ of $1{,}984$ at 3M, all at 6M and 1Y), so the first row differs from Table~\ref{tab:results_full_combined} at 1M and 3M. Mean\,{\scriptsize$\pm$\,std} over 5 runs (observation draw $i$ paired with training subset $i$).}
\label{tab:fixed_k}
\small
\setlength{\tabcolsep}{5pt}
\begin{tabular}{l cccc}
\toprule
Observations & 1M & 3M & 6M & 1Y \\
\midrule
All (Table~\ref{tab:results_full_combined}) & 19.4\pms{0.1} & 27.4\pms{0.1} & 35.2\pms{0.1} & 48.4\pms{0.2} \\
$K{=}4$ & 18.5\pms{0.1} & 20.4\pms{0.1} & 22.3\pms{0.2} & 25.5\pms{0.2} \\
\bottomrule
\end{tabular}
\end{table}

\paragraph{Number of Observations}
To separate the number of observations from the span, we recompute the embeddings with a fixed number $K$ of observations per sensor and fit the linear probe on them (Fig.~\ref{fig:temporal_factors}b). At the 1Y window, mIoU grows by about $7.5$ points per doubling of $K$ up to $K{=}16$ ($17.0$, $24.5$, $32.3$ and $39.7$ for $K{=}2$, $4$, $8$ and $16$) and more slowly beyond ($42.8$, $44.4$ and $46.7$ at $K{=}24$, $32$ and $40$, against $48.0$ for the probe of Table~\ref{tab:results_full_combined} on the same windows). With $40$ dates evenly spaced over the year, as in the sampling of Table~\ref{tab:results_full_combined}, the probe reaches $48.3$: at equal number of observations, spreading them over the whole cycle is worth $1.6$ points over drawing them at random. With $K{=}4$ at every window (Table~\ref{tab:fixed_k}, evaluated on all test windows at 1Y), extending the window from 1M to 1Y still raises mIoU from $18.5$ to $25.5$, whereas at 1Y increasing the observations from $4$ to all those available raises it from $25.5$ to $48.4$. Span and seasonal coverage thus matter on their own, but at the annual scale the dominant factor is how densely the window is sampled, consistent with the accuracy drop reported by \cite{feng2025tessera} below roughly $10$ to $20$ valid observations per year.

\paragraph{Training Budget}
Because the annual label of PASTIS-R supervises every window of a patch, short windows provide more training samples per epoch than 1Y ($17{,}424$ at 1M against $1{,}452$ at 1Y) and therefore more optimizer steps (Table~\ref{tab:train_budget}). To equalize them, we retrain both models with a single window per patch and window length, drawn once before computing the embeddings and kept fixed for all epochs, so that each patch contributes one sample per epoch, as at 1Y. Its start date is uniform over all dates that keep the whole window within the labeled year, so the first and last months of the year are covered less often than by the tiled windows.  This control therefore changes both the number of training steps and the seasonal coverage of the training windows. Evaluated on the same test windows as Table~\ref{tab:results_full_combined}, TESSERA-UNet drops from $35.5$ to $27.0$ mIoU at 1M, from $41.0$ to $31.2$ at 3M and from $45.3$ to $39.2$ at 6M. TESSERA-LP drops from $18.0$ to $16.9$, from $26.7$ to $22.1$ and from $35.2$ to $31.0$ at the same windows. The ranking $\text{1M}<\text{3M}<\text{6M}<\text{1Y}$ is unchanged. The tiled windows therefore favor the short windows, through their additional samples, their full coverage of the year or both, and the gap to 1Y in Table~\ref{tab:results_full_combined} may be underestimated. Since the control changes both factors at once, it does not show how much each of them contributes.

\subsection{Qualitative Results}
\label{subsec:qualitative_results}

\begin{figure*}[t]
    \centering
    \includegraphics[width=1\linewidth]{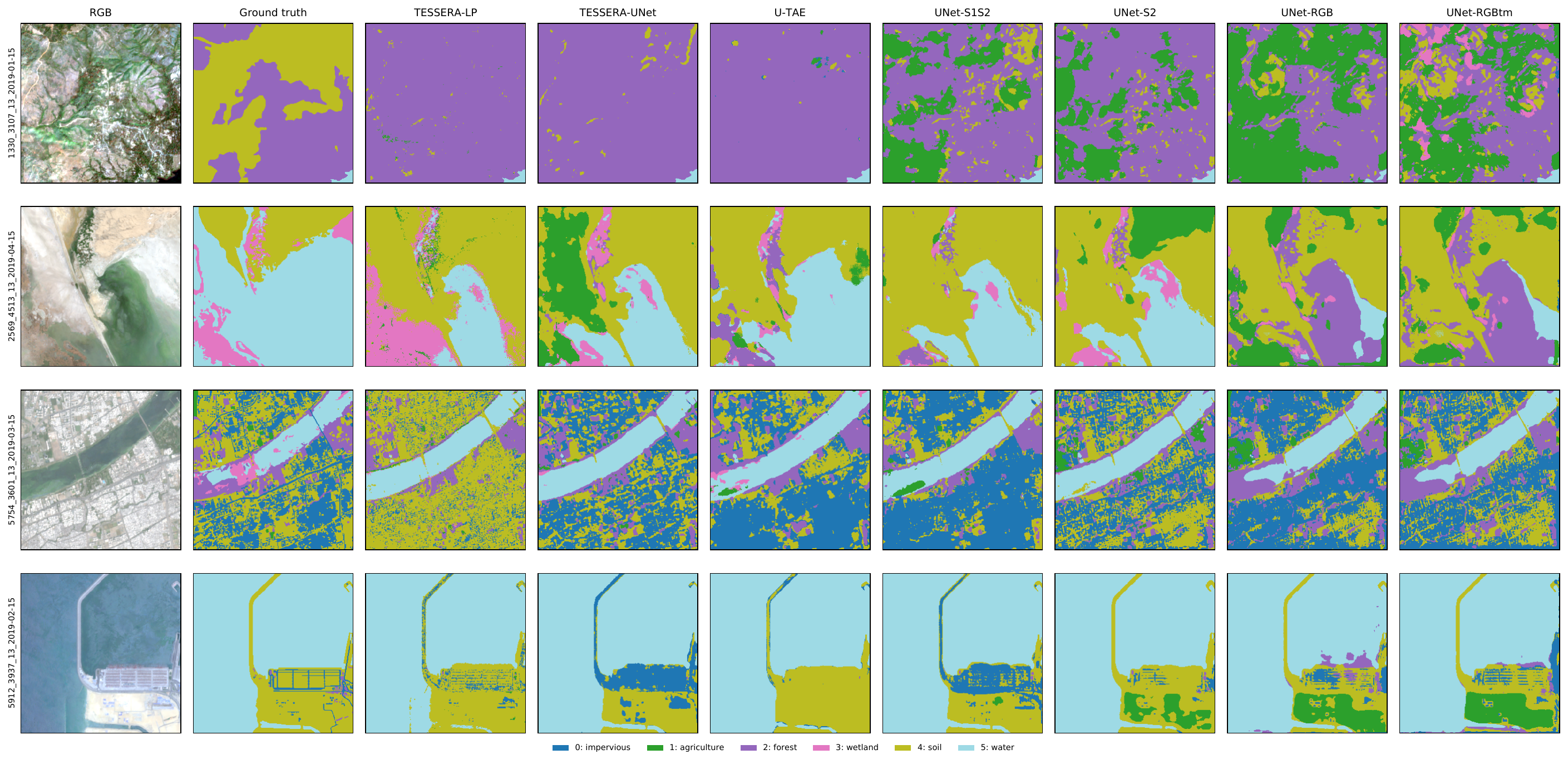}
    \caption{Qualitative results on the DEN test set (1Y window). Each row shows a test sample (RGB composite and ground truth) followed by the predictions of all compared models.}
    \label{fig:qualitative_den}
\end{figure*}
\begin{figure*}[t]
    \centering
    \includegraphics[width=1\linewidth]{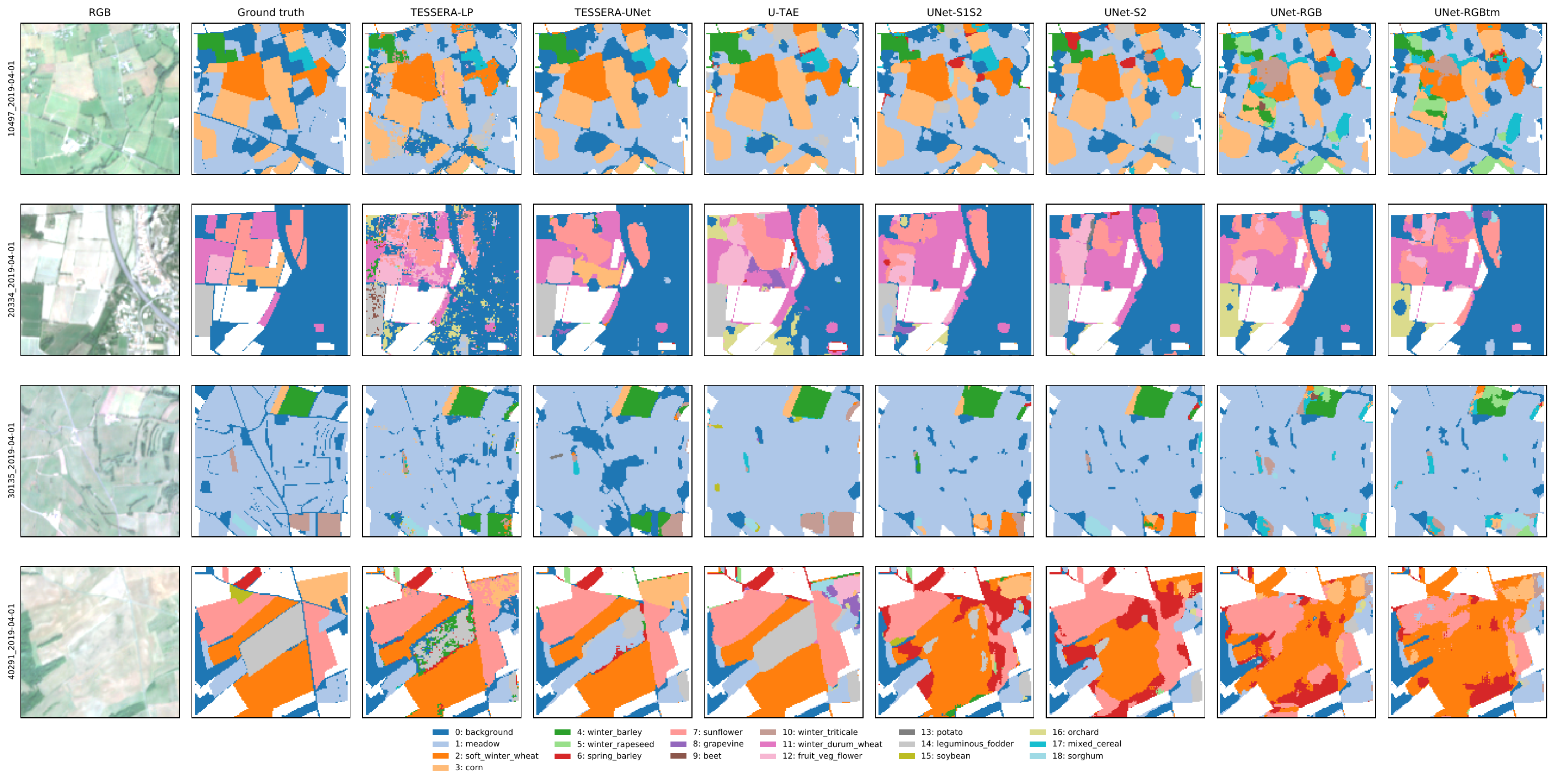}
    \caption{Qualitative results on the PASTIS-R test set (1Y window). Each row shows a test sample (RGB composite and ground truth) followed by the predictions of all compared models.}
    \label{fig:qualitative_pastis}
\end{figure*}

Figures~\ref{fig:qualitative_den} and~\ref{fig:qualitative_pastis} depict examples from all the models presented in Section~\ref{subsubsec:compared_models}. The samples are drawn at random from the test set, without any manual selection. To ease the comparison between the different semantic segmentation masks, we include both the RGB rendered sample and the ground truth.

\subsubsection{Comparison Across Models}
On DEN (Figure~\ref{fig:qualitative_den}), the TESSERA-based predictions are consistently smoother and less speckled than those of the from-scratch UNets, which exhibit clear spectrally driven confusions. The turbid, sediment-laden water of the coastal lagoon (second row) is partially labeled as forest or soil by the RGB variants, and bare-soil areas are contaminated with spurious agriculture speckle (first and last rows). This smoothness, however, occasionally comes at the cost of over-generalization, as in the mountainous scene (first row), where both TESSERA-LP and TESSERA-UNet absorb the soil region into forest. U-TAE produces maps close to those of TESSERA-UNet on DEN, including the same over-generalization in the mountainous scene.

PASTIS-R (Figure~\ref{fig:qualitative_pastis}) visually confirms the complementary failure modes suggested by the quantitative results. The composite-based UNets produce spatially coherent parcels but frequently assign them the wrong crop type, e.g., large spring-barley regions where the ground truth indicates sunflower and winter wheat (last row). They recover the geometry but not the phenology. TESSERA-LP shows the opposite behavior, correctly identifying the crop of most parcels while suffering from salt-and-pepper noise and blurred boundaries, as it operates pixel-wise with no spatial aggregation. U-TAE, which also sees the time series, recovers the crop type of most parcels with coherent boundaries, but misses or confuses some of them, e.g., the winter-barley parcels at the bottom of the third row. TESSERA-UNet combines both sources of information and delivers the maps closest to the ground truth, respecting both parcel boundaries and crop identity.

\subsubsection{Comparison Across Temporal Windows}
\begin{figure*}[tp]
    \centering
    \includegraphics[width=1\linewidth]{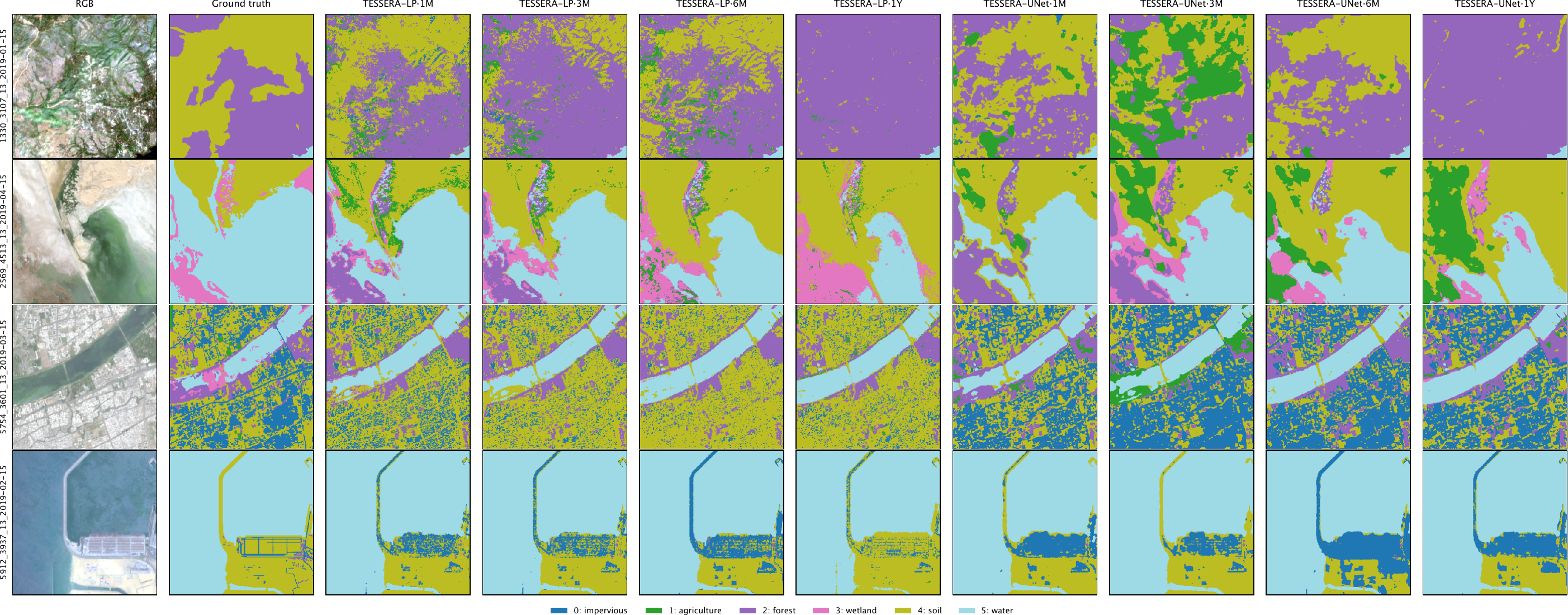}
    \caption{Qualitative evolution across temporal windows on the DEN test set. For each test sample (RGB composite and ground truth), the predictions of TESSERA-LP and TESSERA-UNet are shown at the 1M, 3M, 6M and 1Y windows.}
    \label{fig:temporal_evolution_den}
\end{figure*}

\begin{figure*}[tp]
    \centering
    \includegraphics[width=1\linewidth]{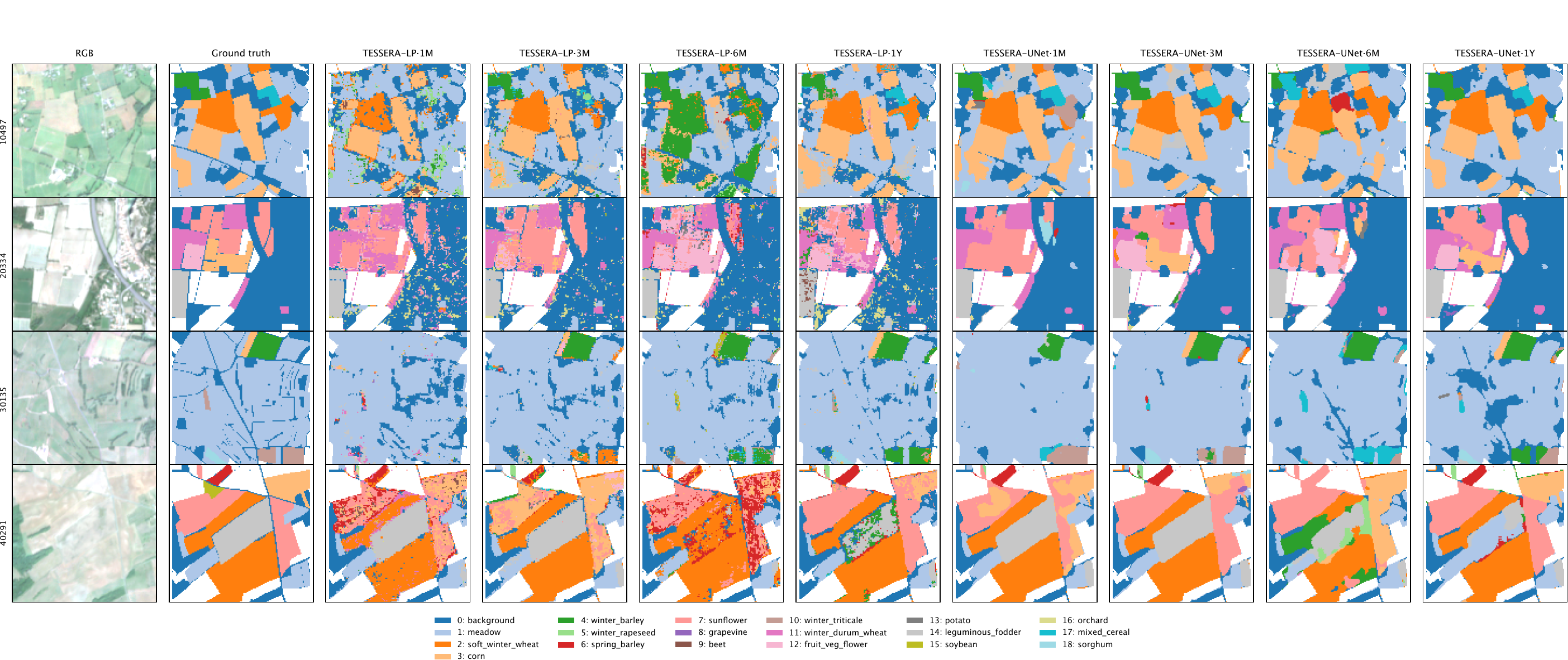}
    \caption{Qualitative evolution across temporal windows on the PASTIS-R test set. For each test sample (RGB composite and ground truth), the predictions of TESSERA-LP and TESSERA-UNet are shown at the 1M, 3M, 6M and 1Y windows.}
    \label{fig:temporal_evolution_pastis}
\end{figure*}

Figures~\ref{fig:temporal_evolution_den} and~\ref{fig:temporal_evolution_pastis} trace how the TESSERA-LP and TESSERA-UNet predictions evolve as the temporal window grows from 1M to 1Y. On PASTIS-R the maps refine substantially. Crop parcels progressively converge to their correct type and the salt-and-pepper noise of the linear probe fades as more of the season is observed, mirroring the steep quantitative gain with window length. On DEN, in contrast, the predictions are already close to their final form at 1M and change little thereafter, consistent with the temporal stability of coarse land-cover classes.


\Needspace*{6\baselineskip}
\section{Discussion}
\label{sec:discussion}
We discuss our results (see Section~\ref{sec:results}) along the main axes that our
evaluation framework analyzes: what the frozen Tessera embeddings are worth for
land-use/land-cover mapping relative to training the same architecture from scratch, and
how that worth evolves as the temporal context from which the embeddings are computed is reduced. On the benchmarks we evaluated, both vary in a way that is consistent with how strongly the classes rely on phenological cues over spectral ones. Below we separate what our results show from how we interpret them.

The task dependence is consistent across our results. On PASTIS-R, where crop types stay spectrally similar for
much of the year and are separated mainly by phenology, the median composites fed to the
from-scratch UNets do not match the performance of the embeddings: our TESSERA-UNet model achieves
$58.3$ mIoU at 1Y, $46\%$ above the best composite-based UNet and $13\%$ above U-TAE, a temporal network trained from scratch, with the highest mIoU at every window. This last margin is an advantage over a temporal baseline trained from scratch under the same protocol. Since U-TAE uses Sentinel-2 only and a different architecture, it does not isolate the contribution of pretraining. On DEN,
whose six coarse classes are spectrally distinct, a single composite already suffices and
the advantage vanishes: TESSERA-UNet sits within ${\sim}2.5\%$ of the strongest from-scratch
UNet (e.g., $43.1$ vs.\ $44.2$ mIoU at 6M), a difference of the order of the seed-to-seed spread
at 1Y, 6M and 3M, and falls slightly behind UNet-S2 at 1M. This equivalence holds only against
full-spectral composites, however: the RGB variants that emulate the common practice of
collapsing Sentinel-2 to three bands trail by $17$--$18\%$ (${\sim}35$ vs.\ $42.9$ mIoU at
1Y).
Interestingly, and in accordance with previous analyses of EO foundation
models~\cite{feng2025tessera}, the
benefit of using embeddings is more evident when labels are scarce: with only $1\%$ of
the labels TESSERA-UNet already achieves ${\sim}24$ mIoU on PASTIS-R, some $50\%$ above
the best composite-based UNet and more than twice U-TAE  (single runs, with two crop classes missing from the $1\%$ subset), and the linear probe outperforms every composite-based UNet at the same label budget, while on DEN the embedding-based models stay essentially
flat from $1\%$ of the labels onwards and the from-scratch models only catch up between
$30\%$ and $100\%$, so the models only draw level on DEN at saturation. In the tasks we evaluated, the embeddings thus pay off when the classes depend on phenology or labels are scarce, and lose their edge only when the classes are spectrally distinct and labels are abundant.

These results support three observations. First, on PASTIS-R the models that keep the time series, U-TAE and TESSERA-UNet, clearly outperform those trained on median composites, whereas on DEN the composite-based UNets are comparable to the embeddings at full supervision. Second, shortening the window costs far more on PASTIS-R than on DEN. Third, within each benchmark, the classes that lose most accuracy at short windows are crops and other vegetation with a marked seasonal cycle, while forest and water are nearly window-invariant.

We interpret these differences as reflecting how strongly the classes depend on their seasonal cycle. Three results support this interpretation independently of the comparison between benchmarks: the class-level pattern is observed within a single benchmark, with geography, observation conditions and annotation protocol held fixed; the same contrast appears without Tessera, since U-TAE gains $13.9$ mIoU over UNet-S2 on the same bands on PASTIS-R at 1Y, whereas on DEN it performs in the range of the composite-based UNets; and, within PASTIS-R, potatoes are not identified by any window that ends before April, whereas meadow is identified by every window (Section~\ref{subsubsec:temporal_factors}). The comparison between benchmarks is, however, also affected by other differences that we do not isolate: PASTIS-R has $18$ fine-grained crop classes (plus background) from parcel declarations over four Sentinel-2 tiles in France, whereas DEN has six coarse classes annotated monthly on $3$\,m imagery at $75$ sites worldwide and resampled to $10$\,m, with different class distributions and observation conditions. Finally, these conclusions concern frozen Tessera V1 embeddings and the three benchmarks evaluated here. Other encoders, tasks or regions may behave differently.

Temporal support, in turn, behaves as a soft and class-dependent constraint rather
than a hard requirement. Contracting the window from one year to one month costs $39\%$
of mIoU on PASTIS-R ($58.3\rightarrow35.5$) but only $5\%$ on DEN
($42.9\rightarrow40.8$). On LUCAS the decline is monotonic yet slow over most of the
range, from $50.8\%$ balanced accuracy at 1Y to $44.6\%$ at 1M, dropping markedly only
under extreme sparsity, to $36.8\%$ from a single day of observation, still $2.9$ times
the $12.5\%$ chance level of balanced accuracy for its eight classes. These margins make low-latency mapping
and fast LULC refresh cycles viable. The cost is also localized, and dataset-level
averages hide it: DEN agriculture loses about half of its 1Y F1 by 1M while forest and
water are almost window-invariant, and on PASTIS-R the most sensitive crops (potatoes,
sorghum and winter triticale) retain only $13$--$22\%$ of their 1Y F1 at 1M against
$84\%$ for meadow. The classes that stay
weak at every window (e.g., DEN wetland, LUCAS water and wetlands) are limited by
under-representation and by small polygons near Sentinel's $10$\,m resolution rather
than by temporal coverage.

The analysis of Section~\ref{subsubsec:temporal_factors} shows that the cost of a short window depends less on its length than on where it falls and how densely it is sampled. On PASTIS-R with TESSERA-UNet, a six-month window covering the growing season matches the full year, and a single month at the peak of the season retains about $90\%$ of it, whereas a single month between October and January retains less than half. At a fixed annual span, accuracy rises steeply with the number of observations up to about $16$ per sensor and more slowly beyond. These results suggest a practical rule for choosing the observation window, within the limits of our three benchmarks: for temporally stable land-cover classes, a single month is close to sufficient; for crop types, the window should cover the growing season of the region rather than a fixed number of months; and in both cases the window should contain enough valid observations, which matters more than its length once the relevant period is covered.

\subsection{Limitations and Future Work}
Our work is not free of limitations, which open avenues for future work.
The proposed evaluation framework varies only the observation window over which a frozen encoder is applied. This encoder was pretrained on full-year sequences and always ingests a fixed-length input, padding by
repetition when fewer distinct acquisitions are available; the degradation we report
is therefore a property of the deployed representation, not evidence that short-window
pretraining would degrade similarly, and training or adapting the encoder with
variable-length windows remains open. Moreover, since the downstream heads are retrained for every window, the framework measures the information retained by the embeddings and not the robustness of a downstream model trained on one window and deployed on another. Evaluating such a train--deployment mismatch is a natural extension. Similarly, our windows are centered on the reference date and use acquisitions up to $W/2$ after it. Strict real-time use would require trailing windows, which we leave for future work. Additionally, the study covers three
LULC benchmarks selected under
the conditions of Section~\ref{sec:datasets} and a single foundation model, although
the framework itself is model-agnostic, applicable to other pixel-wise
embedding products, and not tied to the mapping setting studied here.
On PASTIS-R, published figures serve only as context (Section~\ref{subsubsec:published_context}), since the downstream protocol behind them could not be reproduced. Our DEN
setup evaluates the benchmark as a six-class land-cover segmentation task at
$10\,\mathrm{m}$ and is therefore anchored on from-scratch UNets trained under an
identical protocol to isolate the contribution of the input. LUCAS has no official partition, so
although we follow the balanced-split practice of prior work~\cite{brown2025alphaearth},
no directly comparable published figures exist, and our random split is not spatially independent at the regional scale (Section~\ref{sec:datasets}); a region-held-out evaluation would give a more conservative estimate. Even on PASTIS-R, the implementation
details of the original Tessera result are not fully specified~\cite{feng2025tessera},
so our downstream setup is validated through the ablations of
Section~\ref{subsubsec:ablation_studies} rather than by exact reproduction.
Finally, the
class-level results suggest a concrete direction for operational LULC
systems, namely
class-adaptive temporal windows that spend long observation periods only on the
phenology-driven classes that require them, while spectrally stable classes are mapped
at the fastest available cadence.

\section{Conclusion}
This work introduces a controlled evaluation framework for the temporal sensitivity of a
pixel-wise Earth Observation model in LULC mapping, reconstructing frozen Tessera
embeddings over observation windows ranging from a full year down to a single day and
reading them out with a linear probe and a UNet head on three human-annotated
benchmarks. Two main conclusions follow for the benchmarks evaluated: the value of a temporally-aggregated embedding is
task-dependent, decisive where classes depend on phenology and marginal where
they are spectrally stable and labels abundant; and its degradation under shorter temporal
windows is gradual and concentrated in the classes with a marked seasonal cycle,
so that even single-date embeddings retain some useful semantic information well above chance.

Taken together, our results support treating a temporally-aggregated embedding as a
strong default for operational LULC mapping, and the temporal window as a tunable
cost rather than a fixed prerequisite. Because our evaluation framework needs no retraining of the
encoder and is agnostic to which model produces the embeddings, it can serve as a
standard diagnostic for the pixel-wise foundation models that continue to appear, and
the per-class curves it yields already indicate how much observation each class
inherently needs. More broadly, our results show that pixel-wise EO
foundation models such as Tessera are viable for low-latency applications under
restricted temporal coverage, while motivating further analysis of class-dependent
temporal requirements and inference-time strategies, in which temporal support is
reported as an explicit axis of evaluation.

\appendix[Reproducibility Details]
\subsection{Data Acquisition and Preprocessing}
We download all available Sentinel-1 and Sentinel-2 (Level-2A) observations directly from Microsoft Planetary Computer (MPC) through its STAC interface, applying identical preprocessing across the three datasets (LUCAS 2022, DEN, and PASTIS-R): each region of interest is queried over the target temporal extent, harmonized to a common $10\,\mathrm{m}$ spatial grid, cloud and quality constraints are applied to Sentinel-2, and the resulting multi-temporal stacks are stored in a standardized per-region format.
Tessera inference is then run deterministically on these aligned time series, with fixed sampling and temporal-window settings centered on each reference date, producing georeferenced embedding outputs so that every experiment can be reproduced end-to-end. The same stacks, including the scene-level cloud filter on Sentinel-2, feed both the Tessera encoder and the from-scratch baselines of Appendix~\ref{app:baseline_inputs}; Sentinel-1 is encoded as in the official Tessera preprocessing, with which the encoder was pretrained.

\subsection{Compared-Model Input Construction}
\label{app:baseline_inputs}
The from-scratch UNet baselines (UNet-RGB, UNet-RGB\textsubscript{tm}, UNet-S2 and UNet-S1S2) consume per-window composites built from the same Sentinel-1/2 acquisitions and the same temporal windows used to compute the Tessera embeddings, so that only the input representation differs from TESSERA-UNet. This appendix details their construction for reproducibility.

\subsubsection{Acquisition and Radiometry}
Sentinel-2 Level-2A bottom-of-atmosphere reflectance is retrieved from MPC, keeping scenes with cloud cover below $15\%$, and resampled to the region-of-interest UTM grid at $10\,\mathrm{m}$ by nearest-neighbor interpolation. Ten bands are retained, in the order [B02, B03, B04, B05, B06, B07, B08, B11, B12, B8A], and stored as 16-bit unsigned integer digital numbers (reflectance scaled by $10^4$, nodata $0$). The Scene Classification Layer (SCL) is kept as an auxiliary band for masking. The reflectance scale is not divided out at this stage, as it is later absorbed by the per-channel standardization. Sentinel-1 Radiometric Terrain Corrected (RTC, $\gamma^0$) backscatter is retrieved in IW mode for the VV and VH polarizations on the same $10\,\mathrm{m}$ grid. Each polarization is converted to a logarithmic scale and stored as a 16-bit signed integer, following the official Tessera preprocessing~\cite{feng2025tessera},
\begin{equation}
\begin{gathered}
  d = 20\log_{10}(a), \\
  \mathrm{DN} = \mathrm{clip}\!\left((d+50)\cdot 200,\,0,\,32767\right),
\end{gathered}
\end{equation}
where $a$ is the RTC $\gamma^0$ backscatter as distributed by MPC, in linear power. We apply $20\log_{10}$ rather than the conventional $10\log_{10}$ to follow the Tessera preprocessing, with which the encoder was pretrained and whose normalization statistics assume this scale; $d$ is therefore twice the conventional decibel value. The encoding is invertible as
\begin{equation}
  d = \mathrm{DN}/200 - 50 ,
\end{equation}
so that a digital number of $0$ maps to $d=-50$, i.e., $-25$\,dB in conventional units, and nodata is encoded as $0$. The floor mainly affects VH over open water: among acquisitions with data, $0.26\%$ of VH values on PASTIS-R and $4.5\%$ on DEN lie at the floor ($18.8\%$ and $28.0\%$ over water), against at most $0.05\%$ of VV values. Floor values share the nodata code, as in the official Tessera encoding. Regions without Sentinel-1 coverage have their two radar channels set to zero.

\subsubsection{Windowed Composites}
For each labeled sample and each temporal window $W\in\{\mathrm{1M},\mathrm{3M},\mathrm{6M},\mathrm{1Y}\}$ (that is, $\{30, 91, 182, 365\}$ days) centered on the same reference date as the embeddings, a single composite is formed per modality. For Sentinel-2, acquisitions falling inside the window are SCL-masked by discarding the classes $\{3, 8, 9, 10, 11\}$ (cloud shadow, medium- and high-probability cloud, thin cirrus, and snow/ice), setting those pixels to NaN, after which a per-pixel temporal median is taken independently for each of the ten bands. Pixels with no valid acquisition in the window fall back to the unmasked median, and any residual NaN is set to zero. For Sentinel-1, a per-pixel temporal median of VV and VH is taken over the window without masking, as radar backscatter is insensitive to clouds. The resulting composites are stored in their raw radiometry (Sentinel-2 digital numbers and Sentinel-1 encoded integers), i.e., without normalization; the factor of two in the Sentinel-1 scale is absorbed by the per-channel $z$-score standardization of the baselines. Since the Sentinel-1 median is not masked, zeros, whether from values at the floor or from acquisitions without data at that pixel, enter it as the lowest value; after compositing, VH is $0$ in at most $0.011\%$ of the pixels on PASTIS-R and in $2.3\%$ (1Y) to $5.2\%$ (1M) on DEN, and VV in at most $1.5\%$ (DEN, 1M).

\subsubsection{Per-Variant Inputs and Standardization}
All four variants feed the identical UNet architecture used for TESSERA-UNet, differing only in the number of input channels. Training uses random $128$ (PASTIS-R) or $256$ (DEN) crops with random horizontal and vertical flips, while evaluation runs on the full image padded to a multiple of $32$. UNet-S2 takes the ten Sentinel-2 bands, and UNet-S1S2 additionally appends the two Sentinel-1 channels. Both are per-channel $z$-score standardized using the train-split mean and standard deviation. UNet-RGB takes the B04/B03/B02 bands (in reflectance) and is likewise $z$-score standardized. UNet-RGB\textsubscript{tm} takes the same three bands but is tonemapped offline instead. For each band, global $2$nd and $98$th percentiles are estimated on a sample of train-split pixels, values are clipped to that range, rescaled to $[0,1]$, and gamma-corrected with $\gamma=0.5$. This variant is trained without further $z$-score standardization, since the tonemapping already fixes the dynamic range. All channel statistics, both the $z$-score moments and the tonemapping percentiles, are estimated on the train split only and reused unchanged for validation and test.

\subsection{Additional Results on the Temporal Factors}
\label{app:temporal_factors}
This appendix gives the full results of Section~\ref{subsubsec:temporal_factors}. All values are test metrics (\%) on PASTIS-R, computed as in Table~\ref{tab:results_full_combined}.

\begin{table*}[t]
\centering
\caption{TESSERA-LP and TESSERA-UNet on PASTIS-R scored separately for each window position. Each embedding uses the S1/S2 acquisitions within $\pm\lfloor d/2\rfloor$ days of its reference date ($d = 30, 91, 182, 365$ for 1M, 3M, 6M, 1Y), so the positions tile 1~Oct~2018--30~Sep~2019. The models are those of Table~\ref{tab:results_full_combined} (one model per window length, trained on all its positions); metrics come from one confusion matrix per position; mean\,{\scriptsize$\pm$\,std} over 5 runs. \emph{Pooled} rows merge all positions of a window length and equal Table~\ref{tab:results_full_combined}. Best position per window length in bold.}
\label{tab:window_position}
\scriptsize
\setlength{\tabcolsep}{4pt}
\begin{tabular}{ll cccc cccc}
\toprule
& & \multicolumn{4}{c}{TESSERA-LP} & \multicolumn{4}{c}{TESSERA-UNet} \\
\cmidrule(lr){3-6} \cmidrule(lr){7-10}
W & Position & Acc & B.Acc & F1 & mIoU & Acc & B.Acc & F1 & mIoU \\
\midrule
\multirow{13}{*}{1M}
 & Oct 2018 & 54.5\pms{0.1} & 22.0\pms{0.1} & 19.9\pms{0.1} & 13.0\pms{0.0} & 65.7\pms{0.3} & 38.3\pms{1.3} & 35.4\pms{0.5} & 24.6\pms{0.3} \\
 & Nov 2018 & 50.8\pms{0.1} & 19.8\pms{0.1} & 18.4\pms{0.1} & 11.8\pms{0.1} & 65.6\pms{0.5} & 38.8\pms{1.1} & 35.8\pms{0.5} & 25.0\pms{0.3} \\
 & Dec 2018 & 51.3\pms{0.1} & 22.8\pms{0.1} & 20.4\pms{0.0} & 13.2\pms{0.0} & 66.3\pms{0.4} & 40.1\pms{1.0} & 37.2\pms{0.3} & 26.1\pms{0.2} \\
 & Jan 2019 & 54.6\pms{0.2} & 25.2\pms{0.1} & 22.5\pms{0.1} & 14.7\pms{0.0} & 68.4\pms{0.4} & 42.8\pms{0.9} & 40.0\pms{0.4} & 28.5\pms{0.3} \\
 & Feb 2019 & 59.3\pms{0.1} & 26.2\pms{0.1} & 25.1\pms{0.1} & 16.7\pms{0.0} & 71.0\pms{0.5} & 45.7\pms{1.1} & 42.8\pms{0.4} & 31.1\pms{0.3} \\
 & Mar 2019 & 56.1\pms{0.1} & 28.7\pms{0.1} & 25.5\pms{0.1} & 16.7\pms{0.0} & 71.4\pms{0.4} & 48.5\pms{1.1} & 45.6\pms{0.5} & 33.2\pms{0.3} \\
 & Apr 2019 & 59.0\pms{0.1} & 31.0\pms{0.1} & 28.9\pms{0.2} & 19.3\pms{0.1} & 73.5\pms{0.4} & 55.8\pms{1.1} & 52.4\pms{0.4} & 38.9\pms{0.4} \\
 & May 2019 & 64.8\pms{0.0} & 31.4\pms{0.2} & 30.7\pms{0.1} & 21.7\pms{0.1} & 76.2\pms{0.5} & 61.5\pms{0.8} & 58.0\pms{0.6} & 44.4\pms{0.6} \\
 & Jun 2019 & \textbf{66.4\pms{0.1}} & \textbf{40.4\pms{0.3}} & \textbf{37.8\pms{0.2}} & \textbf{27.1\pms{0.2}} & \textbf{79.4\pms{0.3}} & \textbf{71.8\pms{0.9}} & \textbf{67.6\pms{0.4}} & \textbf{53.7\pms{0.5}} \\
 & Jul 2019 & 64.5\pms{0.1} & 36.5\pms{0.2} & 35.1\pms{0.3} & 24.3\pms{0.2} & 79.1\pms{0.4} & 70.9\pms{0.7} & \textbf{67.6\pms{0.4}} & 53.5\pms{0.4} \\
 & Aug 2019 & 61.9\pms{0.1} & 33.0\pms{0.2} & 30.9\pms{0.1} & 21.0\pms{0.1} & 75.4\pms{0.3} & 60.1\pms{0.9} & 56.8\pms{0.2} & 43.5\pms{0.2} \\
 & Sep 2019 & 60.8\pms{0.1} & 33.1\pms{0.2} & 29.9\pms{0.2} & 20.0\pms{0.2} & 73.3\pms{0.4} & 55.2\pms{1.0} & 52.0\pms{0.2} & 38.9\pms{0.2} \\
 \cmidrule(lr){2-10}
 & \emph{Pooled} & 58.7\pms{0.1} & 29.2\pms{0.1} & 27.5\pms{0.1} & 18.0\pms{0.1} & 72.1\pms{0.4} & 52.5\pms{1.0} & 49.5\pms{0.2} & 35.5\pms{0.2} \\
\midrule
\multirow{5}{*}{3M}
 & Oct--Dec 2018 & 57.8\pms{0.1} & 28.5\pms{0.1} & 25.9\pms{0.1} & 17.1\pms{0.0} & 68.1\pms{0.2} & 41.1\pms{0.7} & 38.8\pms{0.3} & 27.4\pms{0.2} \\
 & Jan--Mar 2019 & 62.1\pms{0.1} & 34.6\pms{0.1} & 32.0\pms{0.0} & 21.7\pms{0.0} & 72.4\pms{0.2} & 50.5\pms{0.5} & 47.7\pms{0.3} & 34.9\pms{0.3} \\
 & Apr--Jun 2019 & \textbf{70.9\pms{0.1}} & \textbf{52.5\pms{0.3}} & \textbf{50.1\pms{0.1}} & \textbf{37.0\pms{0.1}} & \textbf{79.5\pms{0.1}} & \textbf{71.7\pms{0.2}} & \textbf{67.7\pms{0.3}} & \textbf{54.1\pms{0.3}} \\
 & Jul--Sep 2019 & 69.6\pms{0.1} & 51.6\pms{0.2} & 48.6\pms{0.2} & 35.2\pms{0.1} & 79.0\pms{0.1} & 69.8\pms{0.4} & 66.8\pms{0.2} & 52.8\pms{0.2} \\
 \cmidrule(lr){2-10}
 & \emph{Pooled} & 65.1\pms{0.0} & 41.8\pms{0.2} & 39.3\pms{0.1} & 26.7\pms{0.0} & 74.8\pms{0.1} & 58.3\pms{0.4} & 55.7\pms{0.1} & 41.0\pms{0.1} \\
\midrule
\multirow{3}{*}{6M}
 & Oct 2018--Mar 2019 & 66.0\pms{0.1} & 42.3\pms{0.2} & 38.4\pms{0.1} & 26.8\pms{0.1} & 72.5\pms{0.2} & 48.7\pms{0.4} & 46.6\pms{0.4} & 34.3\pms{0.3} \\
 & Apr--Sep 2019 & \textbf{74.9\pms{0.1}} & \textbf{63.6\pms{0.2}} & \textbf{59.9\pms{0.3}} & \textbf{45.9\pms{0.2}} & \textbf{80.7\pms{0.1}} & \textbf{75.4\pms{0.6}} & \textbf{72.2\pms{0.3}} & \textbf{58.6\pms{0.3}} \\
 \cmidrule(lr){2-10}
 & \emph{Pooled} & 70.4\pms{0.1} & 53.0\pms{0.2} & 49.0\pms{0.2} & 35.2\pms{0.1} & 76.6\pms{0.1} & 62.1\pms{0.5} & 60.0\pms{0.3} & 45.3\pms{0.3} \\
\midrule
1Y & Oct 2018--Sep 2019 & 76.1\pms{0.0} & 67.2\pms{0.1} & 62.6\pms{0.2} & 48.4\pms{0.2} & 80.6\pms{0.1} & 74.0\pms{0.4} & 71.9\pms{0.3} & 58.3\pms{0.4} \\
\bottomrule
\end{tabular}
\end{table*}

\begin{figure}[t]
\centering
\includegraphics[width=\columnwidth]{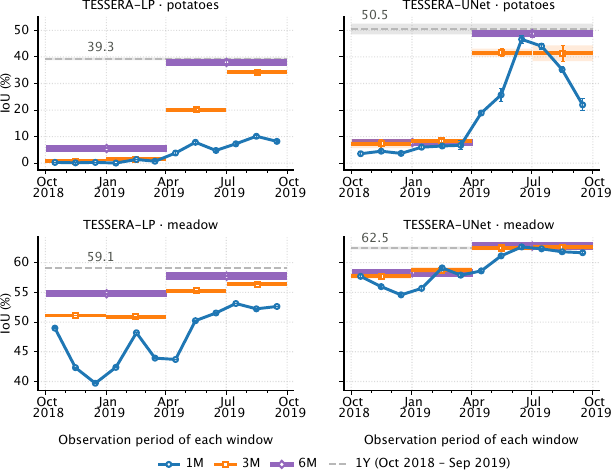}
\caption{Per-class test IoU (\%) on PASTIS-R for each window position, for potatoes (top) and meadow (bottom), the classes that lose most and least F1 between 1Y and 1M in Fig.~\ref{fig:per_class}. Models and protocol as in Table~\ref{tab:window_position}. Each horizontal segment spans the observation period of one window; the dashed line is the 1Y value. Error bars and shaded bands show $\pm1$ std over 5 runs.}
\label{fig:window_position_classes}
\end{figure}

\begin{table}[t]
\centering
\caption{Training budget on PASTIS-R and its effect. Top: training samples per epoch and optimizer steps over 80 epochs (batch $64$) with the tiled windows of Table~\ref{tab:results_full_combined}. Bottom: test mIoU on the test windows of Table~\ref{tab:results_full_combined}, training on the tiled windows (Table~\ref{tab:results_full_combined}) or on a single random window per patch and window length, which gives every window length the samples and steps of 1Y  but covers the first and last months of the year less often than the tiled windows. Random-window results are a single draw.}
\label{tab:train_budget}
\small
\setlength{\tabcolsep}{4pt}
\resizebox{\columnwidth}{!}{%
\begin{tabular}{l cccc}
\toprule
 & 1M & 3M & 6M & 1Y \\
\midrule
Samples per epoch & 17{,}424 & 5{,}808 & 2{,}904 & 1{,}452 \\
Optimizer steps & 21{,}760 & 7{,}200 & 3{,}600 & 1{,}760 \\
\midrule
TESSERA-LP, tiled windows (Table~\ref{tab:results_full_combined}) & 18.0 & 26.7 & 35.2 & 48.4 \\
TESSERA-LP, one random window per patch & 16.9 & 22.1 & 31.0 & (48.4) \\
\midrule
TESSERA-UNet, tiled windows (Table~\ref{tab:results_full_combined}) & 35.5 & 41.0 & 45.3 & 58.3 \\
TESSERA-UNet, one random window per patch & 27.0 & 31.2 & 39.2 & (58.3) \\
\bottomrule
\end{tabular}%
}
\end{table}

\subsubsection{Observations per Window}
\label{app:observations}
Table~\ref{tab:window_obs} reports, for the test samples of each benchmark and window, the number of distinct Sentinel-2 acquisition days and Sentinel-1 passes, how often a sensor has no acquisition in the window (on LUCAS, Sentinel-1 is then filled with zeros; on DEN and PASTIS-R, the closest acquisition outside the window is used), and the offset between the reference date and the annotation date (Section~\ref{sec:framework}).

\begin{table}[t]
\centering
\caption{Observations per window on the test samples. Observations are given as median (25th--75th percentile). The fallback columns give the share of samples without any acquisition of that sensor in the window; in brackets, the maximum distance in days between the replacement acquisition and the reference date $t$. $|t-t_{ann}|$ in days, median (25th--75th percentile) [maximum]; on LUCAS the 1Y window is the calendar year 2022, and on PASTIS-R the label is annual.}
\label{tab:window_obs}
\small
\resizebox{\columnwidth}{!}{%
\begin{tabular}{ll r cc cc c}
\toprule
Dataset & $W$ & $N$ & S2 days & S1 passes & No S2 (\%) & No S1 (\%) & $|t-t_{ann}|$ \\
\midrule
\multirow{6}{*}{LUCAS} & 1D & 13{,}565 & 1 (1--1) & 0 (0--0) & -- & 75.1 & 3 (1--8) [101] \\
 & 1W & 13{,}748 & 1 (1--2) & 2 (1--2) & -- & 2.8 & 3 (1--8) [101] \\
 & 1M & 14{,}050 & 3 (2--5) & 8 (7--10) & -- & 0 & 3 (1--8) [101] \\
 & 3M & 14{,}175 & 8 (5--12) & 24 (21--29) & -- & 0 & 3 (1--8) [101] \\
 & 6M & 14{,}200 & 15 (10--21) & 46 (43--58) & -- & 0 & 3 (1--8) [101] \\
 & 1Y & 14{,}200 & 27 (21--37) & 109 (87--118) & -- & 0 & calendar year \\
\midrule
\multirow{4}{*}{DEN} & 1M & 480 & 3 (1--5) & 5 (3--8) & 18.8 [137] & 3.8 [143] & 14 \\
 & 3M & 480 & 9 (4--14) & 14 (8--25) & 6.7 [137] & 2.3 [143] & 14 \\
 & 6M & 480 & 18 (10--28) & 29 (15--51) & 1.5 [137] & 0.8 [143] & 14 \\
 & 1Y & 480 & 34 (23--50) & 56 (30--106) & 0 & 0 & 14 \\
\midrule
\multirow{4}{*}{PASTIS-R} & 1M & 5{,}952 & 4 (3--6) & 15 (15--20) & 5.9 [42] & 0 & -- \\
 & 3M & 1{,}984 & 12 (7--17) & 46 (43--58) & 0 & 0 & -- \\
 & 6M & 992 & 24 (17--38) & 95 (85--114) & 0 & 0 & -- \\
 & 1Y & 496 & 47 (31--73) & 190 (176--235) & 0 & 0 & -- \\
\bottomrule
\end{tabular}%
}
\end{table}

On DEN, we analyze the impact of using acquisitions outside of the temporal windows by reporting separately results of 1M and 3M on the unscoped set (default behavior) and the scoped one. 
We observe that they are not spread evenly: they are $106$ of the $480$ tile-months at 1M ($90$ without Sentinel-2, $18$ without Sentinel-1) and $43$ at 3M, and at 3M most of them fall between June and September and in three of the $20$ test areas. The 1Y model, which never needs such a replacement, also reaches a higher mIoU on them. Table~\ref{tab:den_borrowed} therefore reports, for the unscoped and scoped set, the mIoU at the short window together with the mIoU of the 1Y model on the same tile-months. With TESSERA-UNet, the two groups reach a similar mIoU at the short window, but the loss with respect to 1Y is larger on the unscoped tile-months: $10.5\%$ against $3.4\%$ at 1M, and $7.3\%$ against none at 3M. With TESSERA-LP, the two groups lose a similar share at 1M ($7.9\%$ and $9.2\%$), whereas at 3M the unscoped tile-months lose more ($13.6\%$ against $5.5\%$). Regarding the differences per land-cover classes, in both groups, forest and water lose at most $2.6$ IoU points with TESSERA-UNet, and the additional loss concentrates on agriculture and soil.

\begin{table}[t]

\centering
\caption{DEN test mIoU (\%) at 1M and 3M for the unscoped (a sensor without acquisitions in the window, filled with its closest acquisition outside it) and scoped (only using acquisition inside the window) tile-months. ``1Y'' is the 1Y model evaluated on the same tile-months, and ``Loss'' the relative drop from 1Y to $W$. Mean\,{\scriptsize$\pm$\,std} over 5 runs; the ``All'' rows are those of Table~\ref{tab:results_full_combined}.}
\label{tab:den_borrowed}
\small
\setlength{\tabcolsep}{4pt}
\resizebox{\columnwidth}{!}{%
\begin{tabular}{ll l r ccc}
\toprule
Model & $W$ & Tile-months & $N$ & mIoU at $W$ & mIoU at 1Y & Loss (\%) \\
\midrule
\multirow{6}{*}{TESSERA-UNet} & \multirow{3}{*}{1M} & Unscoped & 106 & 40.9\pms{0.8} & 45.7\pms{1.9} & 10.5 \\
 & & Scoped & 374 & 40.3\pms{0.1} & 41.7\pms{0.9} & 3.4 \\
 & & All & 480 & 40.8\pms{0.1} & 42.9\pms{1.0} & 4.9 \\
\cmidrule(lr){2-7}
 & \multirow{3}{*}{3M} & Unscoped & 43 & 42.5\pms{1.8} & 45.8\pms{2.0} & 7.3 \\
 & & Scoped & 437 & 42.5\pms{0.1} & 42.3\pms{0.9} & $-0.4$ \\
 & & All & 480 & 42.7\pms{0.2} & 42.9\pms{1.0} & 0.4 \\
\midrule
\multirow{6}{*}{TESSERA-LP} & \multirow{3}{*}{1M} & Unscoped & 106 & 39.3\pms{0.1} & 42.6\pms{0.4} & 7.9 \\
 & & Scoped & 374 & 37.5\pms{0.1} & 41.3\pms{0.1} & 9.2 \\
 & & All & 480 & 37.9\pms{0.1} & 42.0\pms{0.2} & 9.7 \\
\cmidrule(lr){2-7}
 & \multirow{3}{*}{3M} & Unscoped & 43 & 36.7\pms{0.2} & 42.4\pms{0.3} & 13.6 \\
 & & Scoped & 437 & 39.3\pms{0.1} & 41.6\pms{0.2} & 5.5 \\
 & & All & 480 & 39.3\pms{0.1} & 42.0\pms{0.2} & 6.5 \\
\bottomrule
\end{tabular}%
}
\end{table}

\subsection{Evaluation Metrics}
\label{app:metrics}

Evaluation is carried out with pixel-level metrics over the set of classes
$\mathcal{C}$ defined by each benchmark. Let $\mathrm{TP}_c$, $\mathrm{FP}_c$ and $\mathrm{FN}_c$ denote,
respectively, the number of true positives, false positives and false
negatives for class $c$, computed by accumulating the confusion matrix over
all pixels in the test set. We report four complementary metrics: overall
accuracy, balanced accuracy, macro-averaged F1 score, and mean
Intersection-over-Union (mIoU). All metrics are reported as percentages and are the standard choice for reporting performance on the evaluated datasets.

\subsubsection{Overall Accuracy}
The fraction of correctly classified pixels,
\begin{equation}
  \mathrm{Acc} = \frac{\sum_{c\in\mathcal{C}} \mathrm{TP}_c}
                      {\sum_{c\in\mathcal{C}} \left(\mathrm{TP}_c + \mathrm{FN}_c\right)} ,
\end{equation}
which equals the total number of correctly labeled pixels divided by the
total number of pixels. Overall accuracy is dominated by
frequent classes and can therefore be misleading under the strong class
imbalance typical of land-cover and crop-type maps.

\subsubsection{Balanced Accuracy}
To compensate for class imbalance, per-class recall
(true positive rate) is averaged as,
\begin{equation}
\begin{gathered}
  \mathrm{Recall}_c = \frac{\mathrm{TP}_c}{\mathrm{TP}_c + \mathrm{FN}_c} ,
  \\
  \mathrm{BalAcc} = \frac{1}{|\mathcal{C}|} \sum_{c\in\mathcal{C}} \mathrm{Recall}_c ,
\end{gathered}
\end{equation}
so that each class contributes equally regardless of its pixel frequency.

\subsubsection{Macro F1 Score}
For each class, precision and recall are computed, and their harmonic mean is considered to
obtain the per-class F1 score,
\begin{equation}
\begin{gathered}
  \mathrm{Precision}_c = \frac{\mathrm{TP}_c}{\mathrm{TP}_c + \mathrm{FP}_c} ,
  \\
  \mathrm{F1}_c = \frac{2\,\mathrm{Precision}_c \cdot \mathrm{Recall}_c}
                       {\mathrm{Precision}_c + \mathrm{Recall}_c} .
\end{gathered}
\end{equation}
The macro F1 score is the unweighted mean over classes,
\begin{equation}
  \mathrm{F1}_{\text{macro}} = \frac{1}{|\mathcal{C}|} \sum_{c\in\mathcal{C}} \mathrm{F1}_c .
\end{equation}
Like balanced accuracy, the macro average treats all classes equally, but by
combining precision and recall it additionally penalizes over-prediction of
majority classes.

\subsubsection{Mean Intersection-over-Union}
The Jaccard index for class $c$ measures the overlap between the predicted and
ground-truth regions,
\begin{equation}
  \mathrm{IoU}_c = \frac{\mathrm{TP}_c}
                        {\mathrm{TP}_c + \mathrm{FP}_c + \mathrm{FN}_c} ,
\end{equation}
and the mean IoU is the average over all classes,
\begin{equation}
  \mathrm{mIoU} = \frac{1}{|\mathcal{C}|} \sum_{c\in\mathcal{C}} \mathrm{IoU}_c .
\end{equation}
mIoU is the standard primary metric for semantic segmentation, as it jointly
captures both false positives and false negatives in a single per-class
overlap ratio.

\subsubsection{Reporting}
Unless stated otherwise, the per-class averages (balanced accuracy, macro F1
and mIoU) are computed over the classes $\mathcal{C}$ \emph{defined} in each benchmark,
dividing by $|\mathcal{C}|$ irrespective of whether a given class is present in a
particular split. For LUCAS, $|\mathcal{C}| = 8$, corresponding to the Level-1 land-cover
classes; for PASTIS-R, $|\mathcal{C}| = 19$, comprising the background class together with
the $18$ crop classes; and for DEN, $|\mathcal{C}| = 6$, with no background class.

\bibliographystyle{ieeenat_fullname}
\bibliography{main}

\end{document}